\documentclass[10pt,twocolumn,letterpaper]{article}

 \usepackage[pagenumbers]{cvpr} % To force page numbers, e.g. for an arXiv version

\usepackage{leftindex}
\usepackage[ruled,vlined,linesnumbered]{algorithm2e}
\usepackage{booktabs}
\usepackage{multirow}
\usepackage[table,xcdraw]{xcolor}
\usepackage[accsupp]{axessibility}
\usepackage{overpic}
\usepackage{threeparttable}

\definecolor{cvprblue}{rgb}{0.21,0.49,0.74}
\definecolor{codegreen}{rgb}{0.0,0.6,0.0}
\definecolor{codegrey}{RGB}{151,151,151}
\definecolor{Ocean}{RGB}{222,235,246}
\usepackage[pagebackref,breaklinks,colorlinks,allcolors=cvprblue]{hyperref}

\usepackage[ruled,vlined,linesnumbered]{algorithm2e}
\makeatletter
\newcommand{\algorithmfootnote}[2][\footnotesize]{%
  \let\old@algocf@finish\@algocf@finish% Store algorithm finish macro
  \def\@algocf@finish{\old@algocf@finish% Update finish macro to insert "footnote"
    \leavevmode\rlap{\begin{minipage}{\linewidth}
    #1#2
    \end{minipage}}%
  }%
}
\usepackage{indentfirst}
\newif\ifcompileimages
\compileimagestrue
\def\paperID{1784} % *** Enter the Paper ID here
\def\confName{CVPR}
\def\confYear{2025}

\title{S3MOT: Monocular 3D Object Tracking with Selective State Space Model}
\author{Zhuohao Yan\quad
Shaoquan Feng
\quad
Xingxing Li$^{\dagger}$ \quad
Yuxuan Zhou\quad
Chunxi Xia\quad
Shengyu Li
\\
School of Geodesy and Geomatics, Wuhan University, China\\
{\tt\small \texttt{\{yanzhh,shqfeng,xingxingli,yuxuanzhou,xiachunxi,lishengyu\}@whu.edn.cn}}\\
\href{https://github.com/bytepioneerX/s3mot}{https://github.com/bytepioneerX/s3mot}
}

\begin{document}

\twocolumn[{%
\renewcommand\twocolumn[1][]{#1}%
\maketitle
\begin{center}
    \centering
    \captionsetup{type=figure}
\vspace{-15pt}
\begin{overpic}
[width=1\linewidth,]{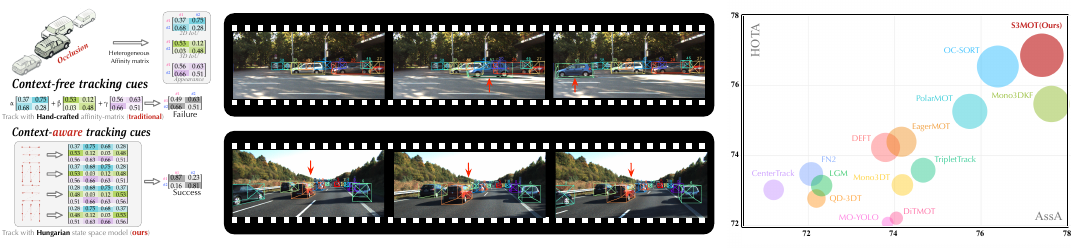}
\put(1,-0.5){\footnotesize{(a)} Context-aware tracking cues}
\put(35,-0.5){\footnotesize{(b)} Qualitative results of S3MOT}
\put(75,-0.5){\footnotesize{(c)} Performance comparison}

\end{overpic}
\captionof{figure}
{(a) contrasts our context-aware tracking cues with traditional context-free methods. Traditional approaches minimize hand-crafted or linearly learned association costs, disregarding interactions between matching pairs. In contrast, our method leverages a four-way scanning mechanism to capture rich contextual knowledge, facilitating efficient information exchange through a global receptive field and dynamic weighting.
(b) presents qualitative results of S3MOT in the challenging scenes. Despite occlusion and fast motion disrupting some tracking cues, our method leverages contextual information for more robust data fusion. Red arrows highlight objects with ambiguous tracking cues.
(c) illustrates the HOTA-AssA comparisons of different trackers. Our monocular 3D tracker S3MOT achieves a new state-of-the-art performance with 76.86~HOTA and 77.41~AssA on the KITTI test benchmark.
}\label{fig1}
\end{center}%
}]

\begin{abstract}
Accurate and reliable multi-object tracking (MOT) in 3D space is essential for advancing robotics and computer vision applications. However, it remains a significant challenge in monocular setups due to the difficulty of mining 3D spatiotemporal associations from 2D video streams. In this work, we present three innovative techniques to enhance the fusion and exploitation of heterogeneous cues for monocular 3D MOT: (1)~we introduce the Hungarian State Space Model (HSSM), a novel data association mechanism that compresses contextual tracking cues across multiple paths, enabling efficient and comprehensive assignment decisions with linear complexity. HSSM features a global receptive field and dynamic weights, in contrast to traditional linear assignment algorithms that rely on hand-crafted association costs. (2)~We propose Fully Convolutional One-stage Embedding (FCOE), which eliminates ROI pooling by directly using dense feature maps for contrastive learning, thus improving object re-identification accuracy under challenging conditions such as varying viewpoints and lighting.  (3)~We enhance 6-DoF pose estimation through VeloSSM, an encoder-decoder architecture that models temporal dependencies in velocity to capture motion dynamics, overcoming the limitations of frame-based 3D inference. Experiments on the KITTI public test benchmark demonstrate the effectiveness of our method, achieving a new state-of-the-art performance of 76.86~HOTA at 31~FPS. Our approach outperforms the previous best by significant margins of +2.63~HOTA and +3.62~AssA, showcasing its robustness and efficiency for monocular 3D MOT tasks. The code and models are available at \url{https://github.com/bytepioneerX/s3mot}.
%\blfootnote{$^\dagger$Corresponding author.}
\end{abstract} 
\section{Introduction}
Visual perception in autonomous systems has commonly relied on LiDAR or multi-sensor arrays to achieve robust 3D multi-object tracking (MOT)~\cite{liu2022bevfusion,pang2023standing,bai2021pointdsc,ding20233dmotformer}. In contrast, monocular 3D MOT~\cite{huang2023delving,li2022time3d,hu2022monocular,zhou2020tracking,marinello2022triplettrack,chaabane2021deft} offers a streamlined and cost-effective solution that extracts 3D tracking information directly from single-camera video streams. Additionally, recent advancements in State Space Models (SSMs), such as S4~\cite{gu2021efficiently}, S4nd~\cite{nguyen2022s4nd}, Mamba~\cite{gu2023mamba}, and SS2D~\cite{liu2024vmamba}, have focused on reducing computational costs while capturing long-range dependencies and preserving high performance. In this paper, we focus on developing monocular 3D MOT algorithms that strike a balance between computational efficiency, robustness, and accurate 3D inference capabilities.

Reliable MOT typically involves three key modules: similarity learning~\cite{pang2021quasi,marinello2022triplettrack,wang2020uncertainty,zhang2021fairmot}, motion modeling~\cite{huang2023delving,zhou2020tracking,weng20203d,hu2019joint,chaabane2021deft,hu2024trackssm}, and data association~\cite{xu2020train,sun2019deep,weng2020gnn3dmot,braso2020learning}. Similarity learning enhances object re-identification and associations, especially for occluded objects. Motion modeling ensures smooth trajectories and reduces ambiguity by utilizing temporal information. Data association integrates various tracking cues (appearance, motion, spatial proximity) to enable robust tracking, typically framed as a graph matching problem that minimizes association costs~\cite{schulter2017deep}.

However, many existing methods focus on isolated feature extraction, struggling to effectively integrate diverse tracking cues, as illustrated in \cref{fig1}(a). For example, some approaches~\cite{hu2022monocular,hu2019joint,wu20213d,wang2022deepfusionmot} rely on affinity matrices with empirically tuned weights, which may not generalize well across different scenarios. Others~\cite{marinello2022triplettrack,sun2019deep,wang2017learning,chu2019famnet,chiu2021probabilistic} use fully convolutional networks (FCNs) or Multilayer Perceptrons (MLPs), limiting their receptive fields and often resulting in suboptimal decisions. Furthermore, Long Short-Term Memory (LSTM)-based models~\cite{xu2020train,tokmakov2021learning,sadeghian2017tracking} encounter issues such as inefficiency and gradient vanishing, which hinder their scalability and make them less suitable for complex, long-sequence tracking tasks.

To address these limitations, we introduce S3MOT, a \textbf{S}elective \textbf{S}tate \textbf{S}pace model-based \textbf{MOT} method that efficiently infers 3D motion and object associations from 2D images through three core components: (i)~Fully Convolutional, One-stage Embedding (FCOE), which uses dense feature maps for contrastive learning to enhance the representational robustness of extracted Re-ID features, mitigating challenges from occlusions and perspective variations; (ii)~VeloSSM, a specialized SSM-based encoder-decoder structure, addresses scale inconsistency and refines motion predictions by modeling temporal dependencies in velocity dynamics; and (iii)~Hungarian State Space Model (HSSM), which employs input-adaptive spatiotemporal scanning and merging, grounded in SSM principles, to associate diverse tracking cues efficiently and ensure reliable tracklet-detection assignments, as illustrated in~\cref{fig1}(b).

By incorporating the SSM, S3MOT effectively captures motion dynamics and integrates diverse tracking cues, greatly enhancing tracking accuracy and robustness, particularly in complex and dynamic environments. Consequently, it achieves more reliable object associations and precise predictions. These strengths are reflected in our exceptional performance on the KITTI public test benchmark, where we achieved 76.86~HOTA and 77.41~AssA, as shown in \cref{fig1}(c). Notably, S3MOT surpasses previous methods such as DEFT~\cite{chaabane2021deft} by +2.63~HOTA and +3.62~AssA and outperforms TripletTrack~\cite{marinello2022triplettrack} by +3.28~HOTA and +2.75~AssA. To the best of our knowledge, this is the first study to apply the selective SSM to monocular 3D MOT, representing a significant advancement toward efficient, scalable, and high-precision tracking.
 
\section{Related Work}

We structure our related work into appearance learning, motion modeling, and data association, focusing on the limitations that motivate our approach.

\textbf{Appearance Learning.}  
Appearance learning plays a crucial role in distinguishing objects across frames, especially in challenging conditions like occlusions, illumination changes, and crowded scenes. Early 3D MOT methods largely focused on simple bounding box tracking and basic motion models, neglecting appearance cues. DeepSORT~\cite{wojke2017simple} improved upon SORT~\cite{bewley2016simple} by incorporating CNN-based appearance features, which enhanced re-identification in complex scenes but still struggled with capturing global dependencies. To address this, approaches like QDTrack~\cite{pang2021quasi} applied contrastive learning to region proposals, achieving better real-time performance. However, these methods remained limited by their inefficiencies in modeling relationships across the entire scene. The shift to more sophisticated models, like Graph Neural Networks (GNNs), brought new advancements. Methods such as GNN3DMOT~\cite{weng2020gnn3dmot} and OGR3MOT~\cite{zaech2022learnable} fused appearance and motion features in 2D and 3D spaces to model object interactions and improve re-identification. Yet, they incurred high computational costs and struggled to fully utilize informative regions, often relying on anchor boxes~\cite{lin2017focal} or RPNs~\cite{ren2016faster}.

\textbf{Motion Modeling.}  
Accurate motion modeling is essential for reliable trajectory prediction in 3D MOT, particularly in dynamic environments. Traditional models like SORT~\cite{bewley2016simple} and DeepSORT~\cite{wojke2017simple} utilized Kalman filters but faced performance drops when dealing with non-linear or unpredictable motion. To improve this, Recurrent Neural Networks (RNNs), including LSTMs, were introduced to capture temporal dependencies, as demonstrated in DEFT~\cite{chaabane2021deft} and QD-3DT~\cite{hu2022monocular}. Despite improvements, these models were hindered by gradient vanishing and scalability issues. Recent attention has turned to State Space Models (SSMs), which offer a more efficient and robust alternative. For example, MambaTrack~\cite{huang2024exploring} replaced Kalman filters with SSMs to better handle non-linear motion in 2D. TrackSSM~\cite{hu2024trackssm} further refined this with a Flow-SSM module, enhancing performance in complex scenarios. However, these advancements have yet to fully address 6-DoF motion modeling in 3D space, leaving a gap that requires more sophisticated techniques.

\textbf{Data Association.}  
Data association remains a core challenge in MOT, often approached as a bipartite graph matching or network flow optimization problem~\cite{zhang2008global}. Early methods relied on hand-crafted cost functions, using empirically tuned weights for features like appearance similarity and spatial distance~\cite{hu2019joint, hu2022monocular}. Although cross-validation has been used to fine-tune these parameters~\cite{wu20213d, feng2024tightly}, the rigidity and subjectivity of such approaches have limited their adaptability. To address these drawbacks, learned cost functions were introduced~\cite{wang2017learning, chiu2021probabilistic}, but these models were often restricted to simple linear combinations. More advanced techniques employed FCNs~\cite{chu2019famnet, marinello2022triplettrack} and RNNs~\cite{xu2020train, sadeghian2017tracking} to construct more complex similarity matrices. However, FCNs are constrained by localized receptive fields, while RNNs suffer from efficiency and scalability issues over long sequences. This underscores the need for more adaptive models that can effectively manage multi-scale and heterogeneous cues in real-time.  
\section{Method}
S3MOT is an end-to-end framework for monocular 3D object detection and tracking. As shown in \cref{fg: architecture}, our method begins with an image input processed by the single-stage detector DD3D~\cite{park2021pseudo} to predict object categories, 2D/3D bounding boxes, and center-ness~\cite{tian2020fcos}. To enhance feature robustness, we introduce the FCOE module, which efficiently extracts Re-ID features from deep feature maps~(\cref{sec:pixel_dense}). We further propose VeloSSM, a data-driven motion model that leverages temporal information to refine tracklet 6-DOF poses~(\cref{sec:deep_motion}). Finally, HSSM combines appearance, motion, and spatial features for robust, consistent object tracking~(\cref{sec:hssm}).

\ifcompileimages
\begin{figure}[!t]
\centering
\includegraphics[width=0.48\textwidth]{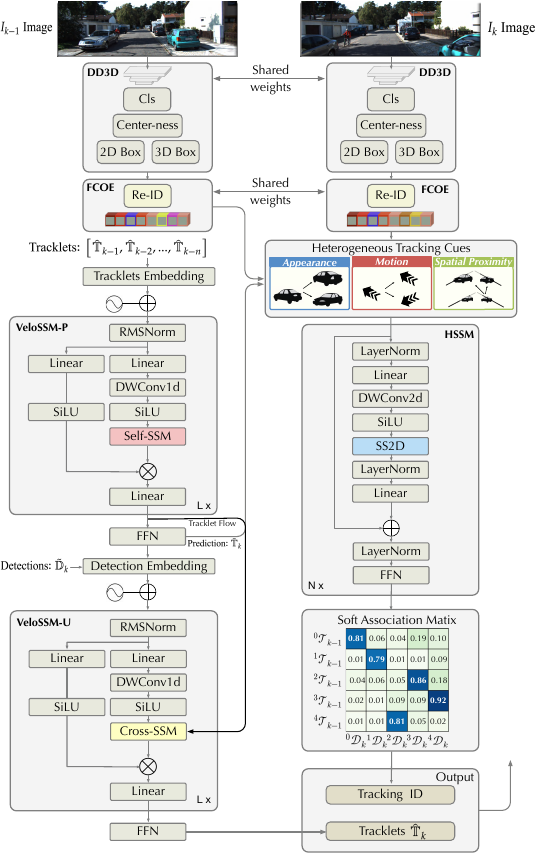}
\caption{Architecture details of S3MOT. The current and historical frame images pass through DD3D to estimate the instance category, 2D bounding box, 3D bounding box, and center-ness. FCOE then extracts Re-ID features from the deep feature map. VeloSSM-P encodes the tracklet flow to predict the current frame's states. HSSM fuses heterogeneous tracking cues to compute a soft association matrix. Finally, VeloSSM-U balances observation and prediction using tracklet flow and confidence, producing the refined tracklet states.}
\label{fg: architecture}
\end{figure}
\else
\textbf{Image compilation is disabled.}
\fi
%-------------------------------------------------------------------------
\subsection{Problem Formulation}
The online MOT problem can be formulated by evaluating the correlations of semantic, spatial, temporal, and other metrics to construct a bipartite matching between the detections $\mathbb{D}$ and the tracks $\mathbb{T}$ in a recursive manner over frames, thereby performing track state optimization or initialization. The detailed definitions of the notations and coordinate systems are provided below.

The states of detected objects are represented as
\begin{equation}
\label{eq:object_state}
\mathbb{D}_k = \left\{\leftindex^j{\boldsymbol{\mathcal{D}}}_{k} \right\}^{M_{k}}_{j=1}
\end{equation}
where $\leftindex^j {\boldsymbol{\mathcal{D}}}_k = \left[ \leftindex^j{\mathbf{q}}_k, \leftindex^j{\mathbf{p}}_k, \leftindex^j{\mathbf{s}}_k, \leftindex^j {c}_k \right]^{\top}$ denotes the state of the $j$-th object among $M_k$ objects at time $t_k$. Specifically, $\leftindex^j{\mathbf{q}}_k$ is the quaternion representing the 3D bounding box pose with full 3-DoF; $\leftindex^j{\mathbf{p}}_k = \left[ \leftindex^j{x}_k, \leftindex^j{y}_k, \leftindex^j{z}_k \right]$ is the 3D position of the object center; $\leftindex^j{\mathbf{s}}_k = \left[ \leftindex^j{w}_k, \leftindex^j{h}_k, \leftindex^j{l}_k \right]$ defines the object’s width, height, and length; and $\leftindex^j{c}_k$ is the confidence score of the 3D bounding box prediction.

The state of tracklets is required for both state prediction and data association. The set of tracklet states is defined as
\begin{equation}
\label{eq:prdictor_state}
\mathbb{T}_k = \left\{ \leftindex^i{\boldsymbol{\mathcal{T}}}_{k}  \right\}^{N_{k}}_{i=1} 
\end{equation}
where $\leftindex^i{\boldsymbol{\mathcal{T}}}_k = \left[ \leftindex^i{\mathbf{q}}_k, \leftindex^i{\mathbf{p}}_k, \leftindex^i{\mathbf{s}}_k, \leftindex^i{\dot{\mathbf{q}}}_k, \leftindex^i{\dot{\mathbf{p}}}_k, \leftindex^i{\dot{\mathbf{s}}}_k \right]^{\top}$ describes the state of the $i$-th tracklet among $N_k$ tracklets at time $t_k$. Here, $\leftindex^i{\dot{\mathbf{q}}}_k$, $\leftindex^i{\dot{\mathbf{p}}}_k$, and $\leftindex^i{\dot{\mathbf{s}}}_k$ represent the 3D angular velocity, linear velocity, and dimension velocity, respectively.

%-------------------------------------------------------------------------
\subsection{Feature-Dense Similarity Learning}
\label{sec:pixel_dense}
Accurate object association across frames demands robust and discriminative feature embeddings. Traditional methods approach similarity learning as a separate post-processing step, relying on sparse anchors~\cite{lu2020retinatrack}, proposals~\cite{pang2021quasi}, or ground-truth boxes~\cite{wojke2017simple}, which limits their effectiveness. To overcome these constraints, we introduce a Fully Convolutional, One-stage Embedding (FCOE) approach, as illustrated in~\cref{fg:pixel-dense}. Unlike prior methods, FCOE operates densely across the feature map, predicting instance embeddings directly without the need for predefined anchors or proposals. This dense setup fully leverages spatial information for training, improving feature learning through contrastive losses.

\ifcompileimages
\begin{figure}[!t]
\centering
\includegraphics[width=0.48\textwidth]{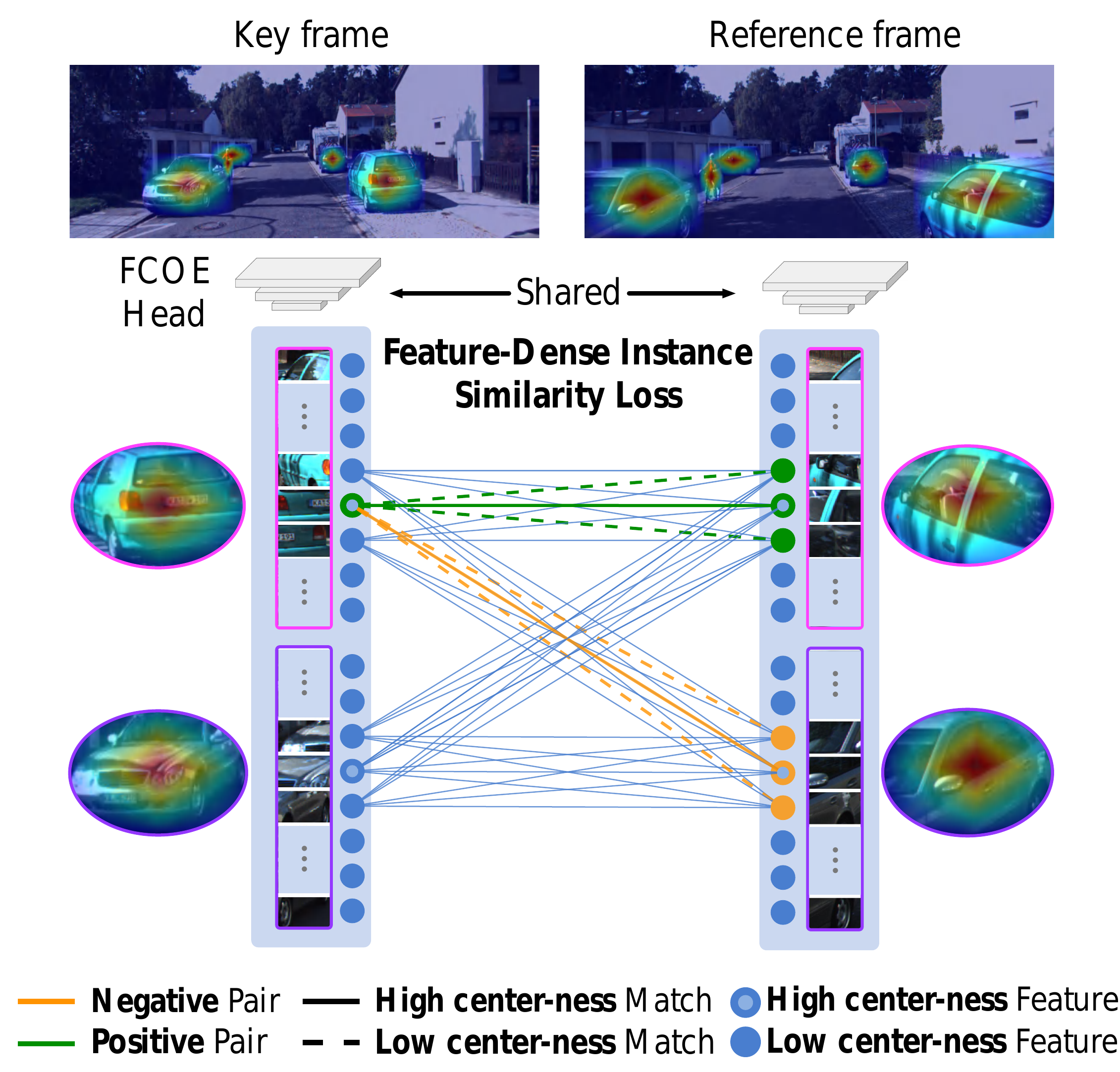}
\caption{Illustration of feature-dense similarity learning. Dense features, comprising high center-ness features (open circles) and low center-ness features (filled circles), are leveraged to construct a discriminative feature space. The feature-dense instance similarity loss operates by comparing dense feature pairs between the keyframe and reference frame, promoting the separation of feature embeddings belonging to different objects while simultaneously pulling embeddings of the same object closer.}
\label{fg:pixel-dense}
\end{figure}
\else
\textbf{Image compilation is disabled.}
\fi

Our method works by computing per-pixel pairwise similarity losses between a keyframe and a reference frame. Specifically, let $F_i \in \mathbb{R}^{H \times W \times C}$ represent the feature maps at layer $i$ of a Feature Pyramid Network (FPN)~\cite{lin2017feature}, with $s$ representing the total stride. Each spatial location $(x, y)$ on $F_i$ maps to a corresponding image location, enabling dense feature learning. A convolutional layer is added to the backbone network to generate 256-D appearance embedding vectors $\mathbf{f}$.

 We label pairs of feature embeddings across frames as positive if they correspond to the same ground-truth object and as negative otherwise. To train these embeddings, we use a cross-entropy loss with non-parametric softmax activation~\cite{wu2018unsupervised}, minimizing the cosine distance for positive pairs and maximizing it for negative pairs,
\begin{equation}
\label{eq:loss_embed_full}
\mathcal{L}_{\text{embed}}= \log \left[ {1+ \leftindex^i{\mathrm{c}}_k  \sum\limits_{\leftindex^i{\mathbf{p}},\leftindex^i{\mathbf{n}}} \exp \left(\leftindex^i{\mathbf{f}}_{k} \cdot \leftindex^i{\mathbf{n}}_{l}-\leftindex^i{\mathbf{f}}_{k} \cdot \leftindex^i{\mathbf{p}}_{l} \right)}  \right]
\end{equation}
where $\leftindex^i{\mathbf{p}}_l$ and $\leftindex^i{\mathbf{n}}_l$ are positive and negative embeddings from the reference frame $l$, and $\leftindex^i{\mathrm{c}}_k$ is the center-ness used in FCOS~\cite{tian2020fcos}. 

To further refine embedding alignment, we apply a cosine similarity loss,
\begin{equation}
\label{eq:loss_cos}
\mathcal{L}_{\text{cosine}}=\left( \frac{\leftindex^i{\mathbf{f}}_{k} \cdot \leftindex^j{\mathbf{f}}_{l}}{\|\leftindex^i{\mathbf{f}}_{k} \cdot \leftindex^j{\mathbf{f}}_{l}\| } -\mathbf{1} \left( i,j \right)  \right)^{2} 
\end{equation}
where $\mathbf{1}(i, j)$ is the indicator function, equating to 1 if $i = j$ and 0 otherwise.

The overall similarity loss for training the FCOE module is defined as,
\begin{equation}
\label{eq:tds_loss}
\mathcal{L}_{\text{FCOE}}=\mathcal{L}_{\text{embed}}+\mathcal{L}_{\text{cosine}}
\end{equation}

Our dense approach ensures that informative regions, both within and outside ground-truth boxes, are effectively utilized, creating a highly discriminative and robust feature space for object association across frames.
%-------------------------------------------------------------------------
\subsection{SSM-based 3D Motion Modeling}
\label{sec:deep_motion}
To infer new object states $\tilde{\mathbb{D}}_k$, we utilize DD3D~\cite{park2021pseudo}, a fully convolutional single-stage 3D detector that directly predicts object states from monocular images. For previously tracked objects, we propose a specialized dual Mamba network, named \textbf{VeloSSM}, designed to model and predict motion flow efficiently. VeloSSM predicts future states $\hat{\mathbb{T}}_k$ while simultaneously refining the current states $\tilde{\mathbb{T}}_k$ by integrating the most recent observations with tracklet motion flow. The overall encoder-decoder architecture is depicted in \cref{fg: architecture}.

\ifcompileimages
\begin{figure*}[!ht]
\centering
\includegraphics[width=\textwidth]{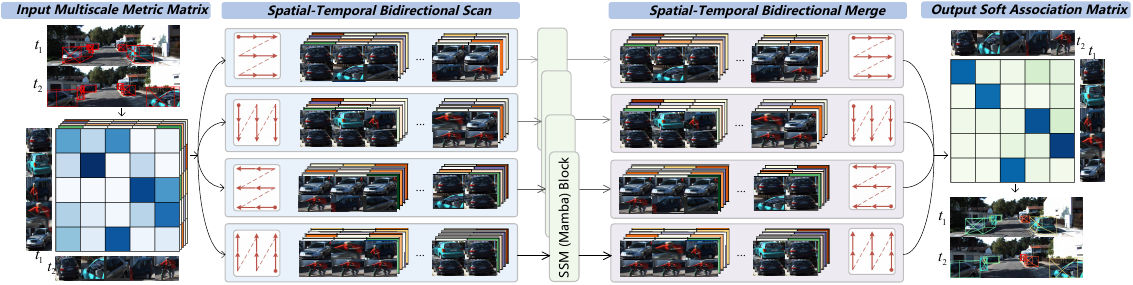}
\caption{Overview of SS2D in our Hungarian State Space Model (HSSM). The input is processed along four separate scan paths, each handled by an independent Mamba block. The outputs from these blocks are then combined via a bidirectional merging process to generate the final 2D feature map.}
\label{fg:hssm}
\end{figure*}
\else
\textbf{Image compilation is disabled.}
\fi

\textbf{Motion Prediction Encoder}.
To capture and utilize temporal coherence, we encode the velocity sequence $\left\{ \left[\dot{\mathbf{q}}_{k-i}^{\mathrm{w}}, \dot{\mathbf{p}}_{k-i}^{\mathrm{w}}, \dot{\mathbf{s}}_{k-i}^{\mathrm{w}} \right] \right\}_{i=1}^{n} \in \mathbb{R}^{n \times 10}$ from $n$ historical frames into a sequence of 320-D embeddings. This sequence is then processed through a Mamba encoder with $L=4$ stacked layers. In this architecture, the step size $\mathbf{\Delta}$, input matrix $\textbf{B}$, and output matrix $\textbf{C}$ (key parameters within the Mamba framework) are dynamically parameterized using the output from the previous layer, a mechanism we term the Self-Selective Scan Mechanism (self-SSM). The encoder generates a 320-D flow feature $\mathcal{F}_k$, which serves two main purposes: (1) predicting future tracklet state $\bar{\mathbb{T}}_k$ through a feed-forward network (FFN) and (2) obtaining the optimized state $\hat{\mathbb{T}}_k$ via the subsequent decoder.

\textbf{Motion Update Decoder}.
To reconcile the observed state $\tilde{\mathbb{D}}_k$ produced by the detector with the predicted state $\bar{\mathbb{T}}_k$ generated by the encoder, we encode the observed and predicted tracklet velocities $\left[\dot{\mathbf{q}}_{k}^{\mathrm{w}}, \dot{\mathbf{p}}_{k}^{\mathrm{w}}, \dot{\mathbf{s}}_{k}^{\mathrm{w}} \right] \in \mathbb{R}^{10}$ into a 320-D embedding, weighted by the 3D detection confidence $c_k$ using a gated MLP. This embedding is then fed into a modified Mamba decoder, comprising a stack of $L=4$ identical layers. Unlike the standard architecture, the parameters $\mathbf{\Delta}$, $\textbf{B}$, and $\textbf{C}$ in this decoder are conditioned on the tracklet flow $\mathcal{F}$ from the encoder, following a cross-selective scanning mechanism (cross-SSM). The output from the final layer is processed through an FFN to yield the refined state $\hat{\mathbb{T}}_k$.

Both the prediction and updating modules are trained using the \textit{disentangled} L1 loss~\cite{simonelli2019disentangling},
\begin{equation}
\label{eq:loss_box3d}
\mathcal{L}_{\text{VeloSSM}}\left( \mathbf{B}^{\ast } ,\hat{\mathbf{B} }  \right) =\frac{1}{8} \left\Vert \mathbf{B}^{\ast } -\hat{\mathbf{B} }  \right\Vert_{1} 
\end{equation} 
where $\mathbf{B}^{\ast }$ and $\hat{\mathbf{B}}$ represent the eight vertices of the ground-truth and the predicted 3D bounding boxes, respectively. The loss is computed iteratively for orientation, position, and dimension components, with non-focused components substituted by ground-truth values.

%-------------------------------------------------------------------------
\subsection{Hungarian State Space Model: HSSM}
\label{sec:hssm}
The Hungarian Algorithm (HA)~\cite{kuhn1955hungarian} is a foundational method for solving the assignment problem in MOT. However, it has two significant limitations in the context of modern deep learning-based tracking systems: (i)~Non-adaptive weighting: HA relies on a predefined cost matrix derived from feature distances but cannot adaptively weight multiple features of different scales or modalities, potentially leading to suboptimal assignments when feature importance varies. (ii)~Non-differentiability: HA involves discrete operations such as permutations and selections, making it unsuitable for integration into end-to-end trainable deep learning frameworks.

To address these limitations, we propose the Hungarian State Space Model (HSSM), a differentiable approximation of the HA that incorporates adaptive feature weighting, as shown in \cref{fg: architecture}. Inspired by SS2D~\cite{liu2024vmamba}, HSSM consists of three key steps: (1) spatiotemporal bidirectional scanning, (2) selective scanning using Mamba blocks, and (3) spatiotemporal bidirectional merging, as shown in \cref{fg:hssm}.

\textbf{Spatial-temporal Bidirectional Scan}.
Let $\mathbf{D} \in \mathbb{R}^{N \times M \times C}$ denote the input distance tensor, where $N$ is the number of existing tracklets, $M$ is the number of current detections, and $C$ represents the number of feature channels or distinct metrics. Each element $\mathbf{D}_{i,j} \in \mathbb{R}^C$ contains the feature distances between track $i$ and detection $j$ across $C$ different feature modalities. To capture both local and global dependencies in $\mathbf{D}$, we define four scanning operations that traverse the tensor in different spatial-temporal directions: (i)~Left-to-Right Horizontal Scan. (ii)~Right-to-Left Horizontal Scan. (iii)~Top-to-Bottom Vertical Scan. (iv)~Bottom-to-Top Vertical Scan. This bidirectional scanning enables the model to capture spatial-temporal context effectively

\textbf{Selective Scanning with Mamba Blocks}.
Each sequence is independently processed in parallel using the Mamba~\cite{gu2023mamba} block, which employs an input-dependent selection mechanism to extract and enhance relevant features while suppressing irrelevant ones. This mechanism enables the model to adaptively weight different feature channels, thereby improving robustness across diverse scenarios.

\textbf{Spatial-temporal Bidirectional Merge}.
After the selective scanning stage, the processed sequences are reshaped to their original spatial dimensions. To convert the aggregated matrix into the soft association matrix $\tilde{\mathbf{A}} \in [0,1]^{N \times M}$, a linear transformation is applied across the feature channels. This transformation is implemented as a point-wise convolution, followed by a sigmoid activation. This operation projects the processed features into a scalar association score for each track-detection pair.

The HA achieves data association by minimizing the total assignment cost of a distance matrix. Our HSSM approximates this by learning a mapping from the distance tensor $\mathbf{D}$ to a soft association matrix $\tilde{\mathbf{A}}$, representing assignment probabilities. By training HSSM to produce $\tilde{\mathbf{A}}$ that closely aligns with ground truth assignments, we enable the model to mimic HA’s behavior while remaining differentiable.

We treat the training of HSSM as a binary classification problem for each track-detection pair, using the ground truth association matrix $\mathbf{A}^\ast \in \{0,1\}^{N \times M}$. To mitigate the class imbalance between matched and unmatched pairs, we apply focal loss~\cite{lin2017focal} with class weighting,
\begin{align}
    \mathcal{L}_\text{HSSM} = - \sum_{i=1}^N \sum_{j=1}^M w_{i,j} \Bigg( & \underbrace{A^\ast_{i,j} (1 - \tilde{A}_{i,j})^\gamma \log(\tilde{A}_{i,j})}_{\text{matched pair term}} \notag \\
    & + \underbrace{(1 - A^\ast_{i,j}) \tilde{A}_{i,j}^\gamma \log(1 - \tilde{A}_{i,j})}_{\text{unmatched pair term}} \Bigg)
\end{align}
where $\gamma$ is the focusing parameter to down-weight easy samples, and $w_{i,j}$ are class weights,
\begin{align}
    w_{i,j} = \begin{cases}
        w_1 = \frac{n_0}{n_1 + n_0}, & \text{if } A^\ast_{i,j} = 1 \\
        w_0 = \frac{n_1}{n_1 + n_0}, & \text{if } A^\ast_{i,j} = 0
    \end{cases}
\end{align}
where $n_1$ and $n_0$ denote the numbers of positive and negative samples, ensuring balanced class contributions during training.
\section{Experiments}
%-------------------------------------------------------------------------
\begin{table*}[!ht]
\caption{Performance comparison of S3MOT with existing methods on the KITTI public testing benchmark. The highest performance values are highlighted in \textbf{\textcolor{red}{red}}, while the second-highest are highlighted in \textbf{\textcolor{blue}{blue}}.}
\centering
\label{tab:KITTI_test}
\setlength{\tabcolsep}{4pt} % Adjust column spacing
\resizebox{\textwidth}{!}{%
\small
\begin{tabular}{l|cc|ccccccccc}
\toprule
Method & Modality& Mode & HOTA$\uparrow$ & DetA$\uparrow$ & AssA$\uparrow$ & DetRe$\uparrow$ & AssRe$\uparrow$ & LocA$\uparrow$ & MOTA$\uparrow$ & FPS $\uparrow$ \\ 
\midrule
Quasi-Dense~\cite{pang2021quasi} $(\textrm{CVPR}^{\prime} 21)$    & Camera & \textcolor{gray}{2D}      & 68.45 & 72.44 & 65.49 & 76.01 & 68.28 & 86.50 & 84.93 & 16 \\
\rowcolor[HTML]{EFEFEF} 
AB3DMOT~\cite{weng20203d} $(\textrm{IROS}^{\prime} 20)$         & LiDAR & 3D       & 69.99 & 71.13 & 69.33 & 75.66 & 72.31 & 86.85 & 83.61 & \textbf{\textcolor{blue}{215}} \\
QD-3DT~\cite{hu2022monocular} $(\textrm{PAMI}^{\prime} 22)$          & Camera  & 3D     & 72.77 & 74.09 & 72.19 & 78.13 & 74.87 & 87.16 & 85.94 & 6 \\
\rowcolor[HTML]{EFEFEF} 
CenterTrack~\cite{zhou2020tracking} $(\textrm{ECCV}^{\prime} 20)$     & Camera & 3D      & 73.02 & 75.62 & 71.20 & 80.10 & 73.84 & 86.52 & \textbf{\textcolor{blue}{88.83}} & 22 \\
Mono3DT~\cite{hu2019joint} $(\textrm{ICCV}^{\prime} 19)$         & Camera & 3D      & 73.16 & 72.73 & 74.18 & 76.51 & 77.18 & 86.88 & 84.28 & 33 \\
\rowcolor[HTML]{EFEFEF} 
TripletTrack~\cite{marinello2022triplettrack} $(\textrm{CVPRW}^{\prime} 22)$    & Camera  & 3D     & 73.58 & 73.18 & 74.66 & 76.18 & 77.31 & 87.37 & 84.32 & -- \\
DEFT~\cite{chaabane2021deft} $(\textrm{CVPR}^{\prime} 21)$            & Camera  & 3D     & 74.23 & 75.33 & 73.79 & 79.96 & 78.30 & 86.14 & 88.38 & 13 \\
\rowcolor[HTML]{EFEFEF} 
EagerMOT~\cite{kim2021eagermot} $(\textrm{ICRA}^{\prime} 21)$         & Camera/LiDAR & 3D & 74.39 & 75.27 & 74.16 & 78.77 & 76.24 & \textbf{\textcolor{blue}{87.17}} & 87.82 & 90 \\
PolarMOT~\cite{kim2022polarmot} $(\textrm{ECCV}^{\prime} 22)$        & LiDAR  & 3D      & 75.16 & 73.94 & \textbf{\textcolor{blue}{76.95}} & \textbf{\textcolor{blue}{80.81}} & 80.00 & 87.12 & 85.08 & 170 \\
\rowcolor[HTML]{EFEFEF} 
OC-SORT~\cite{cao2023observation} $(\textrm{CVPR}^{\prime} 23)$         & Camera  & \textcolor{gray}{2D}     & \textbf{\textcolor{blue}{76.54}} & \textbf{\textcolor{red}{77.25}} & 76.39 & 80.64 & \textbf{\textcolor{blue}{80.33}} & 87.01 & \textbf{\textcolor{red}{90.28}} & \textbf{\textcolor{red}{793}}\\
\hline
S3MOT~$(\textrm{Ours})$               & Camera & 3D      & \textbf{\textcolor{red}{76.86}} & \textbf{\textcolor{blue}{76.95}} & \textbf{\textcolor{red}{77.41}} & \textbf{\textcolor{red}{83.79}} & \textbf{\textcolor{red}{81.01}} & \textbf{\textcolor{red}{87.87}} & 86.93 & 31\\
\bottomrule
\end{tabular}%
}
\end{table*}

\begin{table}[ht]
\caption{Comparison with state-of-the-art 3D MOT trackers on the KITTI validation set. Trackers highlighted in blue utilize the same detector. Bold numbers represent the best performance. $\dag$ indicates the use of Point-RCNN as the detector, while $\ddag$ denotes the use of DD3D. The matching criterion is set to $\text{IoU} = 0.25$.}
\label{tab:kitti-validation}
\resizebox{0.48\textwidth}{!}{%
\begin{tabular}{l|cccc}
\toprule
Method & sAMOTA$\uparrow$ & AMOTA$\uparrow$ & AMOTP$\uparrow$ & MOTP$\uparrow$ \\ \midrule
QD-3DT~\cite{hu2022monocular} & 75.94 & 33.69 & 61.87 & 66.61 \\
AB3DMOT\dag ~\cite{weng20203d} & 86.44 & 41.36 & 77.39 & 80.83 \\
PC3T\dag ~\cite{wu20213d} & 87.08 & 40.50 & 75.18 & \textbf{80.84} \\
\small{DeepFusionMOT\dag ~\cite{wang2022deepfusionmot}} & 90.51 & 44.87 & \textbf{79.47} & 79.83 \\
\midrule
\rowcolor{Ocean} AB3DMOT\ddag ~\cite{weng20203d} & 95.78 & 48.74 & 76.22 & 74.56 \\
\rowcolor{Ocean} PC3T\ddag~\cite{wu20213d} & 96.52 & 49.22 & 75.23 & 73.39 \\
\rowcolor{Ocean} \small{DeepFusionMOT\ddag~\cite{wang2022deepfusionmot}} & 96.68 & 49.49 & 76.39 & 74.66 \\
\hline
\rowcolor{Ocean} S3MOT~(Ours) & \textbf{96.96} & \textbf{49.73} & {77.25} & 73.25 \\ \bottomrule
\end{tabular}
}
\vspace{0.1cm}
\end{table}

\begin{table}[ht]
\centering
\caption{Results on the KITTI validation set using various motion modelings and pairwise costs, where App., IoU, and Mot. represent our proposed appearance cost, IoU-based cost, and motion cost, respectively. Bold numbers denote the best performance. The matching criterion is $\text{IoU} = 0.25$.}
\label{tab:ablation1}
\resizebox{0.47\textwidth}{!}{%
\begin{tabular}{l p{0.5cm} p{0.5cm} p{0.5cm} ccc}
\toprule
\multirow{2}{*}{Method} & \multicolumn{3}{c}{Cost} & \multirow{2}{*}{AMOTA$\uparrow$} & \multirow{2}{*}{AMOTP$\uparrow$} & \multirow{2}{*}{FRAG$\downarrow$} \\ \cmidrule(lr){2-4}
                           & App. & IoU & Mot.       &                                  &                                   &                                   \\ \midrule
KF3D                        & \checkmark & \checkmark & \checkmark & 47.23                          & 76.94                                 & 397                                \\
LSTM                         & \checkmark & \checkmark          & \checkmark & 49.55                          & 77.17                                & 246                                \\ \midrule
\multirow{3}{*}{VeloSSM}      & \checkmark          & -          & -          & 43.54                          & 76.85                                & 966                               \\
                           & \checkmark & -          & \checkmark          & 47.99                          & 76.93                                & 397                                \\
                           & \checkmark & \checkmark          & \checkmark & \textbf{49.73}                 & \textbf{77.25}                        & \textbf{241}                       \\ \bottomrule
\end{tabular}%
}
\end{table}

\begin{table}[ht]
\centering
\caption{Results on the KITTI validation set using various matching algorithms. Bold numbers denote the best performance.}
\label{tab:ablation2}
\resizebox{0.48\textwidth}{!}{%
\begin{tabular}{l|p{0.8cm}|cccc}
\toprule
{Method} & \centering $\text{IoU}$ & {sAMOTA$\uparrow$} & {MOTA$\uparrow$} & {IDS$\downarrow$} & {FRAG$\downarrow$} \\ \midrule
\multirow{2}{*}{\centering Hungarian} & \centering  0.3 & 94.14 & 86.05 & 194 & 274 \\
                                      & \centering 0.5 & 86.69 & 74.77 & 164 & 366 \\ \midrule
\multirow{2}{*}{\centering Greedy}    & \centering 0.3 & 94.05 & 85.91 & 201 & 280 \\
                                      & \centering 0.5 & 86.75 & 74.43 & 173 & 372 \\ \midrule
\multirow{2}{*}{\centering HSSM}      & \centering 0.3 & \textbf{94.32} & \textbf{86.35} & \textbf{178} & \textbf{259} \\
                                      & \centering 0.5 & \textbf{87.24} & \textbf{75.07} & \textbf{150} & \textbf{353} \\ \bottomrule
\end{tabular}
}
\end{table}
%-------------------------------------------------------------------------
\subsection{Datasets and Metrics}
\textbf{Datasets.} The proposed method is evaluated on both the validation and test sets of the KITTI MOT benchmark~\cite{geiger2012we}. Although KITTI provides an official 2D MOT benchmark, its ground truth is not publicly available. Therefore, we report our results by projecting our 3D MOT outputs onto the image plane. For our 3D MOT evaluation, we utilize the KITTI validation set. To ensure consistency with prior work~\cite{hu2022monocular}, we follow the data split protocol, designating sequences 1, 4, 11, 12, 13, 14, 15, and 18 as the validation set, while the remaining sequences are used for training. 

\textbf{Evaluation Metrics.} To comprehensively evaluate the performance of our tracking system, we employ both 3D MOT metrics~\cite{weng20203d} and standard MOT metrics, including HOTA~\cite{luiten2021hota} and CLEAR~\cite{bernardin2008evaluating}. Specifically, the 3D MOT metrics include average MOT accuracy (AMOTA), average MOT precision (AMOTP), and scaled AMOTA (sAMOTA). The HOTA metrics consist of higher-order tracking accuracy (HOTA), detection accuracy (DetA), and association accuracy (AssA), while the CLEAR metrics focus on MOT accuracy (MOTA), MOT precision (MOTP), the number of ID switches (IDSW), the number of trajectory fragmentations (FRAG), frames per second (FPS), and other related measures. By using both 3D-specific and standard MOT evaluation metrics, we aim to provide a comprehensive assessment of the tracking pipeline performance in terms of accuracy, precision, and robustness across different scenarios.
%-------------------------------------------------------------------------
\subsection{Implementation Details}
\textbf{Network Specification.}
We employ the lightweight DLA-34 architecture~\cite{yu2018deep} extended with an FPN as the backbone. The backbone is initialized with parameters pre-trained on 2D object detection using the COCO dataset~\cite{lin2014microsoft} and further pre-trained on dense depth prediction with the DDAD15M dataset.

\textbf{Training.}
For our submissions to the KITTI public benchmark, we trained on the entire detection and tracking dataset, including the validation set. In other experiments, we used only the data outside the validation set for training. The 3D detector and feature-dense similarity learning model were jointly trained with a batch size of 24 images per GPU across 4 Tesla A100 GPUs, resulting in a total batch size of 96. The VeloSSM and HSSM models were jointly trained with a batch size of 128 on a single GeForce RTX 4090 GPU.

\textbf{Inference.}
For the post-processing step involving non-maximum suppression (NMS)~\cite{girshick2014rich}, we set the 2D Intersection over Union (IoU) threshold to 0.75, the 2D confidence threshold to 0.5, and the Bird's Eye View (BEV) IoU threshold to 0.01. During inference, our system is capable of processing multiple scenes simultaneously, using a batch size of 48 on a single GeForce RTX 4090 GPU.

%-------------------------------------------------------------------------
\subsection{Benchmark evaluation}
\textbf{Results on KITTI test set.} The proposed S3MOT surpasses existing monocular 3D MOT methods, achieving 76.86 HOTA at 31 FPS on the official KITTI leaderboard. (see \cref{tab:KITTI_test}). S3MOT excels in association performance with 77.41~AssA, demonstrating its robust ability to maintain accurate and consistent object associations. This result highlights the ability of the model to reliably track objects in dynamic and cluttered environments. Although the DetA of S3MOT at 76.95 slightly trails OC-SORT’s 77.25, it compensates with superior overall tracking performance. The significant improvements in HOTA and AssA highlight the effectiveness of S3MOT in delivering stable and reliable object associations in challenging scenarios. By striking an optimal balance between detection precision and temporal association strength, S3MOT emerges as a comprehensive and reliable solution for monocular 3D MOT.

\textbf{Results on KITTI validation set.}
To thoroughly assess the 3D tracking performance of S3MOT, we conduct experiments on the KITTI validation set and compare it against existing online tracking methods. As presented in ~\cref{tab:kitti-validation}, S3MOT achieves state-of-the-art results in sAMOTA and AMOTA, demonstrating notable gains in tracking accuracy and robustness. 
We further assess performance using Point-RCNN and DD3D-based detectors. Although PC3T benefits from the LiDAR precision of Point-RCNN, S3MOT surpasses PC3T by 9.88\% in sAMOTA and 9.23\% in AMOTA. This highlights the effectiveness of S3MOT in extracting and leveraging discriminative features. While DeepFusionMOT attains higher AMOTP due to multimodal fusion, S3MOT offers a superior balance between accuracy and complexity, with sAMOTA and AMOTA improvements of 6.53\% and 4.86\%. Although localization precision is marginally lower, S3MOT excels in maintaining tracking consistency under occlusions and abrupt motion changes. Notably, even with shared detectors (highlighted in blue), S3MOT achieves the highest overall performance, validating the impact of our architectural innovations and feature learning strategies. These results establish S3MOT as a robust solution for challenging 3D MOT tasks and a new benchmark in the field.
%-------------------------------------------------------------------------
\ifcompileimages
\begin{figure*}
\centering
\includegraphics[width=\textwidth]{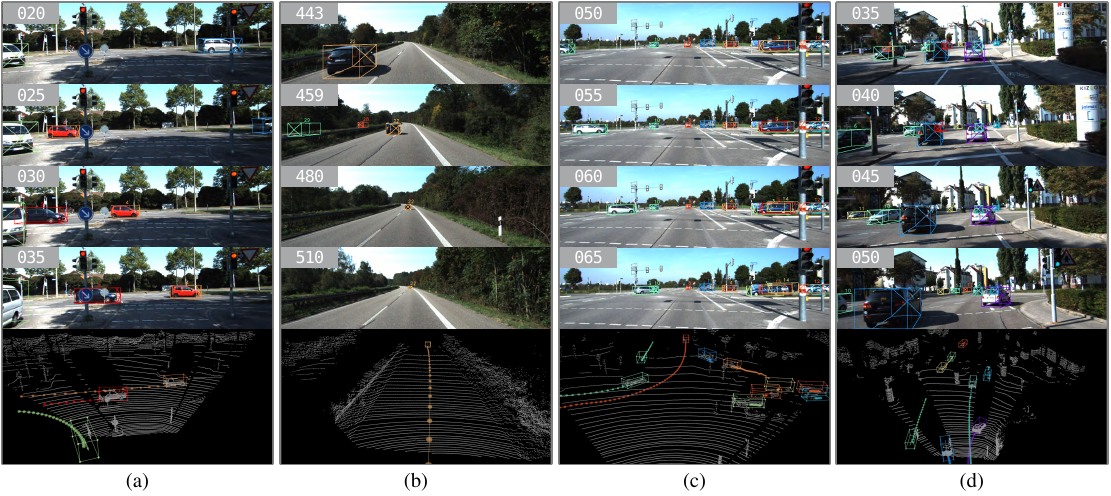}
\caption{Examples of 3D MOT results on the KITTI tracking benchmark. The first four rows show tracked objects projected onto the image plane, with distinct object IDs displayed above each 3D bounding box. In the final row, detected object states for the current frame are represented by bounding boxes in the LiDAR coordinate system, with historical trajectories illustrated using ellipse symbols.}
\label{fg:viskitti}
\end{figure*}
\else
\textbf{Image compilation is disabled.}
\fi

\subsection{Qualitative Results}
We present representative monocular 3D MOT results in \cref{fg:viskitti}, where tracking results are projected onto the image plane for enhanced clarity, and LiDAR point clouds provide spatial reference. The visualizations illustrate four test sequences, presenting 3D bounding boxes with historical trajectories marked by ellipsoid indicators. In sequence (a), our tracker reliably maintains object $\#3$ even amidst significant truncation. Sequence (b) demonstrates robust tracking of the fast-moving, distant object $\#22$, achieved by dynamically adjusting metric weights across scales to address the challenges of monocular 3D tracking. Sequences (c) and (d) further highlight our tracker’s resilience in occlusion-heavy environments, effectively associating current detections with predicted states to ensure continuous and accurate tracking.
%-------------------------------------------------------------------------
\subsection{Ablation Study}
For a fair comparison, all ablation experiments are trained on the training set and tested on the validation set.

\textbf{Heterogeneous Tracking Cues.}
We assess the impact of various tracking cues, including appearance features, IoU-based costs, and motion costs. As shown in \cref{tab:ablation1}, our method, which integrates all three cues, achieves the highest AMOTA and the lowest FRAG, demonstrating the effectiveness of our comprehensive approach. Notably, even when using appearance cues alone, our model achieves a competitive 43.54~AMOTA, illustrating the power of our proposed feature-dense similarity learning. This result highlights the strength of FCOE in utilizing appearance information to maintain robust and accurate tracking, even without relying on additional motion or IoU-based costs.

\textbf{Temporal Motion Modeling.}
We further evaluate the performance of various motion estimators, including KF3D (Kalman Filter), LSTM, and our proposed VeloSSM. As shown in~\cref{tab:ablation1}, VeloSSM consistently outperforms the traditional methods. While KF3D achieves 47.23~AMOTA and LSTM improves this to 49.55~AMOTA, VeloSSM reaches the highest 49.73~AMOTA and the lowest 241~FRAG. These results highlight the superior efficacy of our selective SSM in modeling temporal motion. By more accurately capturing motion dynamics over time, our approach leads to significant improvements in both tracking stability and overall performance.

\textbf{Data Association Methods.}
In the ablation study presented in \cref{tab:ablation2}, we evaluate various data association algorithms across different IoU thresholds, demonstrating the superior performance of our proposed HSSM. The results confirm the robustness of our proposal across all metrics, highlighting its stability, accuracy, and reliability in tracking. By dynamically integrating multiple cues, HSSM achieves more precise global associations, resulting in fewer identity switches and fragmentations and ensuring consistent object continuity, even in complex scenarios.
\section{Conclusion}
%This work introduces a real-time framework for monocular 3D MOT, utilizing the selective SSM architecture to combine appearance, motion, and spatial proximity into a unified soft association matrix with adaptive weights. This integration significantly enhances tracking accuracy and robustness, setting a new benchmark on the KITTI dataset with state-of-the-art monocular 3D tracking performance. We hope this approach will inspire further research on efficient and robust monocular 3D tracking solutions for real-world applications.
This work introduces a real-time framework for monocular 3D MOT, leveraging the selective SSM architecture to combine appearance, motion, and spatial proximity into a unified soft association matrix with adaptive weights. By integrating these diverse tracking cues, the framework significantly improves tracking accuracy and robustness, even in complex and dynamic environments. This novel approach surpasses existing monocular 3D MOT methods, achieving the best results on the KITTI dataset and setting a new standard for tracking performance. We hope that this work will inspire further research into more efficient and robust monocular 3D tracking solutions, particularly in real-world applications where both real-time performance and scalability are critical for practical deployment.

\clearpage
{
    \small
    \bibliographystyle{ieeenat_fullname}
    \bibliography{main}

\begin{thebibliography}{56}
\providecommand{\natexlab}[1]{#1}
\providecommand{\url}[1]{\texttt{#1}}
\expandafter\ifx\csname urlstyle\endcsname\relax
  \providecommand{\doi}[1]{doi: #1}\else
  \providecommand{\doi}{doi: \begingroup \urlstyle{rm}\Url}\fi

\bibitem[Bernardin and Stiefelhagen(2008)]{bernardin2008evaluating}
Keni Bernardin and Rainer Stiefelhagen.
\newblock Evaluating multiple object tracking performance: the clear mot metrics.
\newblock \emph{EURASIP Journal on Image and Video Processing}, 2008:\penalty0 1--10, 2008.

\bibitem[Bewley et~al.(2016)Bewley, Ge, Ott, Ramos, and Upcroft]{bewley2016simple}
Alex Bewley, Zongyuan Ge, Lionel Ott, Fabio Ramos, and Ben Upcroft.
\newblock Simple online and realtime tracking.
\newblock In \emph{2016 IEEE international conference on image processing (ICIP)}, pages 3464--3468. IEEE, 2016.

\bibitem[Bras{\'o} and Leal-Taix{\'e}(2020)]{braso2020learning}
Guillem Bras{\'o} and Laura Leal-Taix{\'e}.
\newblock Learning a neural solver for multiple object tracking.
\newblock In \emph{Proceedings of the IEEE/CVF conference on computer vision and pattern recognition}, pages 6247--6257, 2020.

\bibitem[Cao et~al.(2023)Cao, Pang, Weng, Khirodkar, and Kitani]{cao2023observation}
Jinkun Cao, Jiangmiao Pang, Xinshuo Weng, Rawal Khirodkar, and Kris Kitani.
\newblock Observation-centric sort: Rethinking sort for robust multi-object tracking.
\newblock In \emph{Proceedings of the IEEE/CVF conference on computer vision and pattern recognition}, pages 9686--9696, 2023.

\bibitem[Chaabane et~al.(2021)Chaabane, Zhang, Beveridge, and O'Hara]{chaabane2021deft}
Mohamed Chaabane, Peter Zhang, J~Ross Beveridge, and Stephen O'Hara.
\newblock Deft: Detection embeddings for tracking.
\newblock \emph{arXiv preprint arXiv:2102.02267}, 2021.

\bibitem[Chiu et~al.(2021)Chiu, Li, Ambru{\c{s}}, and Bohg]{chiu2021probabilistic}
Hsu-kuang Chiu, Jie Li, Rare{\c{s}} Ambru{\c{s}}, and Jeannette Bohg.
\newblock Probabilistic 3d multi-modal, multi-object tracking for autonomous driving.
\newblock In \emph{2021 IEEE international conference on robotics and automation (ICRA)}, pages 14227--14233. IEEE, 2021.

\bibitem[Chu and Ling(2019)]{chu2019famnet}
Peng Chu and Haibin Ling.
\newblock Famnet: Joint learning of feature, affinity and multi-dimensional assignment for online multiple object tracking.
\newblock In \emph{Proceedings of the IEEE/CVF International Conference on Computer Vision}, pages 6172--6181, 2019.

\bibitem[Ding et~al.(2023)Ding, Rehder, Schneider, Cordts, and Gall]{ding20233dmotformer}
Shuxiao Ding, Eike Rehder, Lukas Schneider, Marius Cordts, and Juergen Gall.
\newblock 3dmotformer: Graph transformer for online 3d multi-object tracking.
\newblock In \emph{Proceedings of the IEEE/CVF International Conference on Computer Vision}, pages 9784--9794, 2023.

\bibitem[Feng et~al.(2024)Feng, Li, Yan, Xia, Li, Wang, and Zhou]{feng2024tightly}
Shaoquan Feng, Xingxing Li, Zhuohao Yan, Chunxi Xia, Shengyu Li, Xuanbing Wang, and Yuxuan Zhou.
\newblock Tightly coupled integration of lidar and vision for 3d multiobject tracking.
\newblock \emph{IEEE Transactions on Intelligent Vehicles}, 2024.

\bibitem[Geiger et~al.(2012)Geiger, Lenz, and Urtasun]{geiger2012we}
Andreas Geiger, Philip Lenz, and Raquel Urtasun.
\newblock Are we ready for autonomous driving? the kitti vision benchmark suite.
\newblock In \emph{2012 IEEE conference on computer vision and pattern recognition}, pages 3354--3361. IEEE, 2012.

\bibitem[Girshick et~al.(2014)Girshick, Donahue, Darrell, and Malik]{girshick2014rich}
Ross Girshick, Jeff Donahue, Trevor Darrell, and Jitendra Malik.
\newblock Rich feature hierarchies for accurate object detection and semantic segmentation.
\newblock In \emph{Proceedings of the IEEE conference on computer vision and pattern recognition}, pages 580--587, 2014.

\bibitem[Gu and Dao(2023)]{gu2023mamba}
Albert Gu and Tri Dao.
\newblock Mamba: Linear-time sequence modeling with selective state spaces.
\newblock \emph{arXiv preprint arXiv:2312.00752}, 2023.

\bibitem[Gu et~al.(2021)Gu, Goel, and R{\'e}]{gu2021efficiently}
Albert Gu, Karan Goel, and Christopher R{\'e}.
\newblock Efficiently modeling long sequences with structured state spaces.
\newblock \emph{arXiv preprint arXiv:2111.00396}, 2021.

\bibitem[Hu et~al.(2024)Hu, Luo, Liu, Wang, and Liu]{hu2024trackssm}
Bin Hu, Run Luo, Zelin Liu, Cheng Wang, and Wenyu Liu.
\newblock Trackssm: A general motion predictor by state-space model.
\newblock \emph{arXiv preprint arXiv:2409.00487}, 2024.

\bibitem[Hu et~al.(2019)Hu, Cai, Wang, Lin, Sun, Krahenbuhl, Darrell, and Yu]{hu2019joint}
Hou-Ning Hu, Qi-Zhi Cai, Dequan Wang, Ji Lin, Min Sun, Philipp Krahenbuhl, Trevor Darrell, and Fisher Yu.
\newblock Joint monocular 3d vehicle detection and tracking.
\newblock In \emph{Proceedings of the IEEE/CVF International Conference on Computer Vision}, pages 5390--5399, 2019.

\bibitem[Hu et~al.(2022)Hu, Yang, Fischer, Darrell, Yu, and Sun]{hu2022monocular}
Hou-Ning Hu, Yung-Hsu Yang, Tobias Fischer, Trevor Darrell, Fisher Yu, and Min Sun.
\newblock Monocular quasi-dense 3d object tracking.
\newblock \emph{IEEE Transactions on Pattern Analysis and Machine Intelligence}, 45\penalty0 (2):\penalty0 1992--2008, 2022.

\bibitem[Huang et~al.(2024)Huang, Yang, Chai, Jiang, and Hwang]{huang2024exploring}
Hsiang-Wei Huang, Cheng-Yen Yang, Wenhao Chai, Zhongyu Jiang, and Jenq-Neng Hwang.
\newblock Exploring learning-based motion models in multi-object tracking.
\newblock \emph{arXiv preprint arXiv:2403.10826}, 2024.

\bibitem[Huang et~al.(2023)Huang, Yang, and Tsai]{huang2023delving}
Kuan-Chih Huang, Ming-Hsuan Yang, and Yi-Hsuan Tsai.
\newblock Delving into motion-aware matching for monocular 3d object tracking.
\newblock In \emph{Proceedings of the IEEE/CVF International Conference on Computer Vision}, pages 6909--6918, 2023.

\bibitem[Kim et~al.(2021)Kim, O{\v{s}}ep, and Leal-Taix{\'e}]{kim2021eagermot}
Aleksandr Kim, Aljo{\v{s}}a O{\v{s}}ep, and Laura Leal-Taix{\'e}.
\newblock Eagermot: 3d multi-object tracking via sensor fusion.
\newblock In \emph{2021 IEEE International conference on Robotics and Automation (ICRA)}, pages 11315--11321. IEEE, 2021.

\bibitem[Kim et~al.(2022)Kim, Bras{\'o}, O{\v{s}}ep, and Leal-Taix{\'e}]{kim2022polarmot}
Aleksandr Kim, Guillem Bras{\'o}, Aljo{\v{s}}a O{\v{s}}ep, and Laura Leal-Taix{\'e}.
\newblock Polarmot: How far can geometric relations take us in 3d multi-object tracking?
\newblock In \emph{European Conference on Computer Vision}, pages 41--58. Springer, 2022.

\bibitem[Kuhn(1955)]{kuhn1955hungarian}
Harold~W Kuhn.
\newblock The hungarian method for the assignment problem.
\newblock \emph{Naval research logistics quarterly}, 2\penalty0 (1-2):\penalty0 83--97, 1955.

\bibitem[Li and Jin(2022)]{li2022time3d}
Peixuan Li and Jieyu Jin.
\newblock Time3d: End-to-end joint monocular 3d object detection and tracking for autonomous driving.
\newblock In \emph{Proceedings of the IEEE/CVF conference on computer vision and pattern recognition}, pages 3885--3894, 2022.

\bibitem[Lin(2017)]{lin2017focal}
T Lin.
\newblock Focal loss for dense object detection.
\newblock \emph{arXiv preprint arXiv:1708.02002}, 2017.

\bibitem[Lin et~al.(2014)Lin, Maire, Belongie, Hays, Perona, Ramanan, Doll{\'a}r, and Zitnick]{lin2014microsoft}
Tsung-Yi Lin, Michael Maire, Serge Belongie, James Hays, Pietro Perona, Deva Ramanan, Piotr Doll{\'a}r, and C~Lawrence Zitnick.
\newblock Microsoft coco: Common objects in context.
\newblock In \emph{Computer Vision--ECCV 2014: 13th European Conference, Zurich, Switzerland, September 6-12, 2014, Proceedings, Part V 13}, pages 740--755. Springer, 2014.

\bibitem[Lin et~al.(2017)Lin, Doll{\'a}r, Girshick, He, Hariharan, and Belongie]{lin2017feature}
Tsung-Yi Lin, Piotr Doll{\'a}r, Ross Girshick, Kaiming He, Bharath Hariharan, and Serge Belongie.
\newblock Feature pyramid networks for object detection.
\newblock In \emph{Proceedings of the IEEE conference on computer vision and pattern recognition}, pages 2117--2125, 2017.

\bibitem[Liu et~al.(2024)Liu, Tian, Zhao, Yu, Xie, Wang, Ye, and Liu]{liu2024vmamba}
Yue Liu, Yunjie Tian, Yuzhong Zhao, Hongtian Yu, Lingxi Xie, Yaowei Wang, Qixiang Ye, and Yunfan Liu.
\newblock Vmamba: Visual state space model.
\newblock \emph{arXiv preprint arXiv:2401.10166}, 2024.

\bibitem[Liu et~al.(2023)Liu, Tang, Amini, Yang, Mao, Rus, and Han]{liu2022bevfusion}
Zhijian Liu, Haotian Tang, Alexander Amini, Xingyu Yang, Huizi Mao, Daniela Rus, and Song Han.
\newblock Bevfusion: Multi-task multi-sensor fusion with unified bird's-eye view representation.
\newblock In \emph{IEEE International Conference on Robotics and Automation (ICRA)}, 2023.

\bibitem[Lu et~al.(2020)Lu, Rathod, Votel, and Huang]{lu2020retinatrack}
Zhichao Lu, Vivek Rathod, Ronny Votel, and Jonathan Huang.
\newblock Retinatrack: Online single stage joint detection and tracking.
\newblock In \emph{Proceedings of the IEEE/CVF conference on computer vision and pattern recognition}, pages 14668--14678, 2020.

\bibitem[Luiten et~al.(2021)Luiten, Osep, Dendorfer, Torr, Geiger, Leal-Taix{\'e}, and Leibe]{luiten2021hota}
Jonathon Luiten, Aljosa Osep, Patrick Dendorfer, Philip Torr, Andreas Geiger, Laura Leal-Taix{\'e}, and Bastian Leibe.
\newblock Hota: A higher order metric for evaluating multi-object tracking.
\newblock \emph{International journal of computer vision}, 129:\penalty0 548--578, 2021.

\bibitem[Marinello et~al.(2022)Marinello, Proesmans, and Van~Gool]{marinello2022triplettrack}
Nicola Marinello, Marc Proesmans, and Luc Van~Gool.
\newblock Triplettrack: 3d object tracking using triplet embeddings and lstm.
\newblock In \emph{Proceedings of the IEEE/CVF Conference on Computer Vision and Pattern Recognition}, pages 4500--4510, 2022.

\bibitem[Nguyen et~al.(2022)Nguyen, Goel, Gu, Downs, Shah, Dao, Baccus, and R{\'e}]{nguyen2022s4nd}
Eric Nguyen, Karan Goel, Albert Gu, Gordon Downs, Preey Shah, Tri Dao, Stephen Baccus, and Christopher R{\'e}.
\newblock S4nd: Modeling images and videos as multidimensional signals with state spaces.
\newblock \emph{Advances in neural information processing systems}, 35:\penalty0 2846--2861, 2022.

\bibitem[Pang et~al.(2021)Pang, Qiu, Li, Chen, Li, Darrell, and Yu]{pang2021quasi}
Jiangmiao Pang, Linlu Qiu, Xia Li, Haofeng Chen, Qi Li, Trevor Darrell, and Fisher Yu.
\newblock Quasi-dense similarity learning for multiple object tracking.
\newblock In \emph{Proceedings of the IEEE/CVF conference on computer vision and pattern recognition}, pages 164--173, 2021.

\bibitem[Pang et~al.(2023)Pang, Li, Tokmakov, Chen, Zagoruyko, and Wang]{pang2023standing}
Ziqi Pang, Jie Li, Pavel Tokmakov, Dian Chen, Sergey Zagoruyko, and Yu-Xiong Wang.
\newblock Standing between past and future: Spatio-temporal modeling for multi-camera 3d multi-object tracking.
\newblock In \emph{Proceedings of the IEEE/CVF Conference on Computer Vision and Pattern Recognition}, 2023.

\bibitem[Park et~al.(2021)Park, Ambrus, Guizilini, Li, and Gaidon]{park2021pseudo}
Dennis Park, Rares Ambrus, Vitor Guizilini, Jie Li, and Adrien Gaidon.
\newblock Is pseudo-lidar needed for monocular 3d object detection?
\newblock In \emph{Proceedings of the IEEE/CVF International Conference on Computer Vision}, pages 3142--3152, 2021.

\bibitem[Ren et~al.(2016)Ren, He, Girshick, and Sun]{ren2016faster}
Shaoqing Ren, Kaiming He, Ross Girshick, and Jian Sun.
\newblock Faster r-cnn: Towards real-time object detection with region proposal networks.
\newblock \emph{IEEE transactions on pattern analysis and machine intelligence}, 39\penalty0 (6):\penalty0 1137--1149, 2016.

\bibitem[Sadeghian et~al.(2017)Sadeghian, Alahi, and Savarese]{sadeghian2017tracking}
Amir Sadeghian, Alexandre Alahi, and Silvio Savarese.
\newblock Tracking the untrackable: Learning to track multiple cues with long-term dependencies.
\newblock In \emph{Proceedings of the IEEE international conference on computer vision}, pages 300--311, 2017.

\bibitem[Schulter et~al.(2017)Schulter, Vernaza, Choi, and Chandraker]{schulter2017deep}
Samuel Schulter, Paul Vernaza, Wongun Choi, and Manmohan Chandraker.
\newblock Deep network flow for multi-object tracking.
\newblock In \emph{Proceedings of the IEEE Conference on Computer Vision and Pattern Recognition}, pages 6951--6960, 2017.

\bibitem[Simonelli et~al.(2019)Simonelli, Bulo, Porzi, L{\'o}pez-Antequera, and Kontschieder]{simonelli2019disentangling}
Andrea Simonelli, Samuel~Rota Bulo, Lorenzo Porzi, Manuel L{\'o}pez-Antequera, and Peter Kontschieder.
\newblock Disentangling monocular 3d object detection.
\newblock In \emph{Proceedings of the IEEE/CVF International Conference on Computer Vision}, pages 1991--1999, 2019.

\bibitem[Sun et~al.(2019)Sun, Akhtar, Song, Mian, and Shah]{sun2019deep}
ShiJie Sun, Naveed Akhtar, HuanSheng Song, Ajmal Mian, and Mubarak Shah.
\newblock Deep affinity network for multiple object tracking.
\newblock \emph{IEEE transactions on pattern analysis and machine intelligence}, 43\penalty0 (1):\penalty0 104--119, 2019.

\bibitem[Tian et~al.(2020)Tian, Shen, Chen, and He]{tian2020fcos}
Zhi Tian, Chunhua Shen, Hao Chen, and Tong He.
\newblock Fcos: A simple and strong anchor-free object detector.
\newblock \emph{IEEE transactions on pattern analysis and machine intelligence}, 44\penalty0 (4):\penalty0 1922--1933, 2020.

\bibitem[Tokmakov et~al.(2021)Tokmakov, Li, Burgard, and Gaidon]{tokmakov2021learning}
Pavel Tokmakov, Jie Li, Wolfram Burgard, and Adrien Gaidon.
\newblock Learning to track with object permanence.
\newblock In \emph{Proceedings of the IEEE/CVF International Conference on Computer Vision}, pages 10860--10869, 2021.

\bibitem[Wang et~al.(2020)Wang, Ancha, Chen, and Held]{wang2020uncertainty}
Jianren Wang, Siddharth Ancha, Yi-Ting Chen, and David Held.
\newblock Uncertainty-aware self-supervised 3d data association.
\newblock In \emph{2020 IEEE/RSJ International Conference on Intelligent Robots and Systems (IROS)}, pages 8125--8132. IEEE, 2020.

\bibitem[Wang and Fowlkes(2017)]{wang2017learning}
Shaofei Wang and Charless~C Fowlkes.
\newblock Learning optimal parameters for multi-target tracking with contextual interactions.
\newblock \emph{International journal of computer vision}, 122\penalty0 (3):\penalty0 484--501, 2017.

\bibitem[Wang et~al.(2022)Wang, Fu, Li, Lai, and He]{wang2022deepfusionmot}
Xiyang Wang, Chunyun Fu, Zhankun Li, Ying Lai, and Jiawei He.
\newblock Deepfusionmot: A 3d multi-object tracking framework based on camera-lidar fusion with deep association.
\newblock \emph{IEEE Robotics and Automation Letters}, 7\penalty0 (3):\penalty0 8260--8267, 2022.

\bibitem[Weng et~al.(2020{\natexlab{a}})Weng, Wang, Held, and Kitani]{weng20203d}
Xinshuo Weng, Jianren Wang, David Held, and Kris Kitani.
\newblock 3d multi-object tracking: A baseline and new evaluation metrics.
\newblock In \emph{2020 IEEE/RSJ International Conference on Intelligent Robots and Systems (IROS)}, pages 10359--10366. IEEE, 2020{\natexlab{a}}.

\bibitem[Weng et~al.(2020{\natexlab{b}})Weng, Wang, Man, and Kitani]{weng2020gnn3dmot}
Xinshuo Weng, Yongxin Wang, Yunze Man, and Kris~M Kitani.
\newblock Gnn3dmot: Graph neural network for 3d multi-object tracking with 2d-3d multi-feature learning.
\newblock In \emph{Proceedings of the IEEE/CVF Conference on Computer Vision and Pattern Recognition}, pages 6499--6508, 2020{\natexlab{b}}.

\bibitem[Wojke et~al.(2017)Wojke, Bewley, and Paulus]{wojke2017simple}
Nicolai Wojke, Alex Bewley, and Dietrich Paulus.
\newblock Simple online and realtime tracking with a deep association metric.
\newblock In \emph{2017 IEEE international conference on image processing (ICIP)}, pages 3645--3649. IEEE, 2017.

\bibitem[Wu et~al.(2021)Wu, Han, Wen, Li, and Wang]{wu20213d}
Hai Wu, Wenkai Han, Chenglu Wen, Xin Li, and Cheng Wang.
\newblock 3d multi-object tracking in point clouds based on prediction confidence-guided data association.
\newblock \emph{IEEE Transactions on Intelligent Transportation Systems}, 23\penalty0 (6):\penalty0 5668--5677, 2021.

\bibitem[Wu et~al.(2018)Wu, Xiong, Yu, and Lin]{wu2018unsupervised}
Zhirong Wu, Yuanjun Xiong, Stella~X Yu, and Dahua Lin.
\newblock Unsupervised feature learning via non-parametric instance discrimination.
\newblock In \emph{Proceedings of the IEEE conference on computer vision and pattern recognition}, pages 3733--3742, 2018.

\bibitem[Xu et~al.(2020)Xu, Osep, Ban, Horaud, Leal-Taix{\'e}, and Alameda-Pineda]{xu2020train}
Yihong Xu, Aljosa Osep, Yutong Ban, Radu Horaud, Laura Leal-Taix{\'e}, and Xavier Alameda-Pineda.
\newblock How to train your deep multi-object tracker.
\newblock In \emph{Proceedings of the IEEE/CVF conference on computer vision and pattern recognition}, pages 6787--6796, 2020.

\bibitem[Xuyang~Bai and Tai(2022)]{bai2021pointdsc}
Xinge Zhu Qingqiu Huang Yilun Chen Hongbo~Fu Xuyang~Bai, Zeyu~Hu and Chiew-Lan Tai.
\newblock {TransFusion}: {R}obust {L}idar-{C}amera {F}usion for {3}d {O}bject {D}etection with {T}ransformers.
\newblock \emph{CVPR}, 2022.

\bibitem[Yu et~al.(2018)Yu, Wang, Shelhamer, and Darrell]{yu2018deep}
Fisher Yu, Dequan Wang, Evan Shelhamer, and Trevor Darrell.
\newblock Deep layer aggregation.
\newblock In \emph{Proceedings of the IEEE conference on computer vision and pattern recognition}, pages 2403--2412, 2018.

\bibitem[Zaech et~al.(2022)Zaech, Liniger, Dai, Danelljan, and Van~Gool]{zaech2022learnable}
Jan-Nico Zaech, Alexander Liniger, Dengxin Dai, Martin Danelljan, and Luc Van~Gool.
\newblock Learnable online graph representations for 3d multi-object tracking.
\newblock \emph{IEEE Robotics and Automation Letters}, 7\penalty0 (2):\penalty0 5103--5110, 2022.

\bibitem[Zhang et~al.(2008)Zhang, Li, and Nevatia]{zhang2008global}
Li Zhang, Yuan Li, and Ramakant Nevatia.
\newblock Global data association for multi-object tracking using network flows.
\newblock In \emph{2008 IEEE conference on computer vision and pattern recognition}, pages 1--8. IEEE, 2008.

\bibitem[Zhang et~al.(2021)Zhang, Wang, Wang, Zeng, and Liu]{zhang2021fairmot}
Yifu Zhang, Chunyu Wang, Xinggang Wang, Wenjun Zeng, and Wenyu Liu.
\newblock Fairmot: On the fairness of detection and re-identification in multiple object tracking.
\newblock \emph{International journal of computer vision}, 129:\penalty0 3069--3087, 2021.

\bibitem[Zhou et~al.(2020)Zhou, Koltun, and Kr{\"a}henb{\"u}hl]{zhou2020tracking}
Xingyi Zhou, Vladlen Koltun, and Philipp Kr{\"a}henb{\"u}hl.
\newblock Tracking objects as points.
\newblock In \emph{European conference on computer vision}, pages 474--490. Springer, 2020.

\end{thebibliography}
}

\clearpage

\appendix
\setcounter{page}{12}
\maketitlesupplementary

\section{Preliminary}
\label{sec:preliminary}
State Space Models (SSMs) represent a class of models that utilize hidden states for sequential autoregressive data modeling. One of the most prominent SSMs is the S4 model~\cite{gu2021efficiently}, a continuous Linear Time Invariant (LTI) system that transforms a 1-D input signal $\mathbf{U}(t)$ into an N-D output signal $\mathbf{y}(t)$. The system dynamics are expressed as follows,
\begin{equation}  
\begin{aligned} 
\label{vanillaSSMs}
\dot{\mathbf{X}}(t)  &= \mathbf{A}(t) \mathbf{X}(t) + \mathbf{B}(t) \mathbf{U}(t) \\  
\mathbf{y}(t)        &= \mathbf{C}(t) \mathbf{X}(t) + \mathbf{D}(t) \mathbf{U}(t)  
\end{aligned}  
\end{equation}
where $\mathbf{X}(t) \in \mathbb{R}^n$ represents the hidden state, $\mathbf{y}(t) \in \mathbb{R}^q$ is the output vector, and $\mathbf{U}(t) \in \mathbb{R}^p$ denotes the input vector. The term $\dot{\mathbf{X}}(t) = \frac{d}{dt}\mathbf{X}(t)$ denotes the derivative of the hidden state. The matrices $\mathbf{A}(t) \in \mathbb{R}^{n \times n}$, $\mathbf{B}(t) \in \mathbb{R}^{n \times p}$, $\mathbf{C}(t) \in \mathbb{R}^{q \times n}$, and $\mathbf{D}(t) \in \mathbb{R}^{q \times p}$ correspond to the state matrix, input matrix, output matrix, and feed-forward matrix, respectively. In the absence of direct feedthrough, $\mathbf{D}(t)$ becomes a zero matrix, simplifying the system to:
\begin{equation}  
\begin{aligned} 
\label{vanillaSSMs_simple}
\dot{\textbf{X}}(t)  &= \textbf{A}(t) \textbf{X}(t) + \textbf{B}(t) \textbf{U}(t) \\  
\textbf{y}(t)        &= \textbf{C}(t) \textbf{X}(t)
\end{aligned}  
\end{equation} 

Recently,~\cite{gu2023mamba} introduced the Mamba module, a data-driven SSM layer. Mamba's key innovation lies in parameterizing the $\mathbf{\Delta}$, $\mathbf{B}$, and $\mathbf{C}$ matrices based on sequence data representations. This mechanism enables the model to selectively focus on or suppress specific sequence state inputs, improving its capacity to capture sequence features and handle long-sequence modeling. Notably, the Mamba module scales linearly with sequence length, maintaining low memory usage and high inference efficiency. Given that the system operates continuously, discretization is required for digital processing. The discretized system using zero-order hold (ZOH) is given by, 
\begin{equation}  
\begin{aligned} 
\label{Mamba_discretize}
\mathbf{X}_t  &= \overline{\mathbf{A}} \mathbf{X}_{t-1} + \overline{\textbf{B}} \mathbf{U}_{t} \\  
\mathbf{y}_t  &= \mathbf{C} \mathbf{X}_t
\end{aligned}  
\end{equation}
where $\overline{\mathbf{A}} = \exp(\mathbf{\Delta} \mathbf{A})$, $\overline{\mathbf{B}} = (\mathbf{\Delta} \mathbf{A})^{-1}(\exp(\mathbf{\Delta} \mathbf{A}) - \mathbf{I}) \cdot \mathbf{\Delta} \mathbf{B}$, and $\mathbf{\Delta}$ is the step size.
%-------------------------------------------------------------------------
\section{Pseudo-code of VeloSSM and HSSM}
\label{sec:pseudo-code}
For clarity, we present the pseudocodes for VeloSSM-P (see~\cref{algo:velossm-p}), VeloSSM-U (see~\cref{algo:velossm-u}), and HSSM (see~\cref{algo:hssm}) individually, though they are trained jointly. The symbol $\odot$ represents the Hadamard product, $\otimes$ denotes tensor multiplication, while $\boxplus$ represents the addition on a manifold.

The distance matrix $\mathbf{D}$ in~\cref{algo:hssm} is computed based on four criteria: 
(1) the appearance similarity \(\mathbf{D}_{\text{App.}}(\mathbf{f}_{\mathbb{T}}, \mathbf{f}_{\mathbb{D}})\), which measures the similarity between the accumulated feature \(\mathbf{f}_{\mathbb{T}}\) of the current tracklets and the learned feature \(\mathbf{f}_{\mathbb{D}}\) from detected objects;  
(2) the overlap of bounding boxes \(\mathbf{D}_{\text{IoU}}(\mathbf{B}_{\mathbb{T}}, \mathbf{B}_{\mathbb{D}})\), which calculates the intersection over union between the temporal projection of the 3D bounding box \(\mathbf{B}_{\mathbb{T}}\) of the tracklets and the detected bounding box \(\mathbf{B}_{\mathbb{D}}\);  
(3) the motion similarity \(\mathbf{D}_{\text{mot.}}(\mathbf{v}_{\mathbb{T}}, \mathbf{v}_{\mathbb{D}})\), which quantifies the similarity between the observed motion vector \(\mathbf{v}_{\mathbb{T}}\) of the tracklets and the pseudo-motion vector \(\mathbf{v}_{\mathbb{D}}\) of the detection;  
(4) the category consistency \(\mathbf{D}_{\text{cls.}}(\boldsymbol{\phi}_{\mathbb{T}}, \boldsymbol{\phi}_{\mathbb{D}})\), which  measures the alignment between the predicted category \(\boldsymbol{\phi}_{\mathbb{T}}\) of the current tracklets and the detected category \(\boldsymbol{\phi}_{\mathbb{D}}\).

\begin{algorithm}[!t]
\SetAlgoLined
\DontPrintSemicolon
\SetNoFillComment
\footnotesize
\KwIn{tracklet state sequence $\hat{\mathbb{T}}_{k-1:k-n}$: \textcolor{codegreen}{$(\mathrm{B}, \mathrm{n}, \mathrm{10})$};}
\KwOut{tracklet state prediction $\bar{\mathbb{T}}_{k}$: \textcolor{codegreen}{$(\mathrm{B}, \mathrm{10})$}; \\ \hspace{1.0cm} tracklet flow $\mathcal{F}_{k-1}$: \textcolor{codegreen}{$(\mathrm{B}, \mathrm{d})$}}
\textcolor{codegrey}{\tcc{tracklet velocity embedding}}
$\mathbb{T}^{\prime}_{k-1:k-n}$: \textcolor{codegreen}{$(\mathrm{B}, \mathrm{n}, \mathrm{10})$} $\leftarrow \text{Difference}(\hat{\mathbb{T}}_{k-1:k-n})$ \;
$\mathbf{T}^{(0)}_{k-1:k-n}$: \textcolor{codegreen}{$(\mathrm{B}, \mathrm{n}, \mathrm{d})$} $\leftarrow \text{Linear}_{\mathbf{T}^{(0)}}(\mathbb{T}^{\prime}_{k-1:k-n})$ \;
\textcolor{codegrey}{\tcc{processing each layer of $L$ layers}}
\ForEach{$\ell \in \left[0, L \right)$}{
\textcolor{codegrey}{\tcc{normalize the input sequence}}
${\hat{\mathbf{T}}}^{(\ell)}_{k-1:k-n}$: \textcolor{codegreen}{$(\mathrm{B}, \mathrm{n}, \mathrm{d})$} $\leftarrow \text{RMSNorm}(\mathbf{T}^{(\ell)}_{k-1:k-n})$\;
\textcolor{codegrey}{\tcc{input projection}}
$\mathbf{x}$: \textcolor{codegreen}{$(\mathrm{B}, \mathrm{n}, \mathrm{D})$} $\leftarrow \text{SiLU}(\text{Conv1d}(\text{Linear}_{\mathbf{x}}({\hat{\mathbf{T}}}^{(\ell)}_{k-1:k-n})))$\;
$\mathbf{z}$: \textcolor{codegreen}{$(\mathrm{B}, \mathrm{n}, \mathrm{D})$} $\leftarrow \text{Linear}_{\mathbf{z}}({\hat{\mathbf{T}}}^{(\ell)}_{k-1:k-n})$\;
$\mathbf{A}$: \textcolor{codegreen}{$(\mathrm{D}, \mathrm{N})$} $\leftarrow  \text{Parameter}$ \;
$\mathbf{B}$: \textcolor{codegreen}{$(\mathrm{B}, \mathrm{n}, \mathrm{N})$} $\leftarrow  \text{Linear}_{\mathbf{B}}(\mathbf{x})$ \;
$\mathbf{C}$: \textcolor{codegreen}{$(\mathrm{B}, \mathrm{n}, \mathrm{N})$} $\leftarrow  \text{Linear}_{\mathbf{C}}(\mathbf{x})$ \;
\textcolor{codegrey}{\tcc{softplus ensures positive $\mathbf{\Delta}$}}
$\mathbf{\Delta}$: \textcolor{codegreen}{$(\mathrm{B}, \mathrm{n}, \mathrm{D})$} $\leftarrow \text{softplus}(\text{Broadcast}_{\mathrm{D}}(\text{Linear}_{\mathbf{\Delta}}(\mathbf{\mathbf{x}})))$ \;
\textcolor{codegrey}{\tcc{discretize}}
$\overline{\mathbf{A}}$: \textcolor{codegreen}{$(\mathrm{B}, \mathrm{n}, \mathrm{D}, \mathrm{N})$} $\leftarrow  \text{exp}(\mathbf{\Delta} \otimes  \mathbf{A})$ \;
$\overline{\mathbf{B}}$: \textcolor{codegreen}{$(\mathrm{B}, \mathrm{n}, \mathrm{D}, \mathrm{N})$} $\leftarrow \mathbf{\Delta} \otimes  \mathbf{B}$ \;
\textcolor{codegrey}{\tcc{Self-SSM}}
$\mathbf{y}$: \textcolor{codegreen}{$(\mathrm{B}, \mathrm{n}, \mathrm{D})$} $\leftarrow \textbf{SSM}(\overline{\mathbf{A}}, \overline{\mathbf{B}}, \mathbf{C})(\mathbf{x})$\;
\textcolor{codegrey}{\tcc{gated output}}
$\mathbf{y}$: \textcolor{codegreen}{$(\mathrm{B}, \mathrm{n}, \mathrm{D})$} $\leftarrow \mathbf{y} \odot \text{SiLU}(\mathbf{z})$\;
\textcolor{codegrey}{\tcc{residual connection}}
$\mathbf{T}^{(\ell +1)}_{k-1:k-n}$: \textcolor{codegreen}{$(\mathrm{B}, \mathrm{n}, \mathrm{d})$} $\leftarrow \text{Linear}_{\mathbf{T}}(\mathbf{y})+\mathbf{T}^{(\ell)}_{k-1:k-n}$ \;
}
$\boldsymbol{w}_{k-1:k-n}$: \textcolor{codegreen}{$(\mathrm{B},\mathrm{n}, \mathrm{10})$} $\leftarrow \text{softmax}(\text{Linear}_{\boldsymbol{w}}(\mathbf{T}^{(L)}_{k-1:k-n}))$ \;
$\mathbb{T}^{\prime}_{k}$: \textcolor{codegreen}{$(\mathrm{B}, \mathrm{10})$} $\leftarrow \text{sum}(\boldsymbol{w}_{k-1:k-n} \odot \mathbb{T}^{\prime}_{k-1:k-n})$ \;
$\hat{\mathbb{T}}_{k}$: \textcolor{codegreen}{$(\mathrm{B}, \mathrm{10})$} $\leftarrow \hat{\mathbb{T}}_{k-1} \boxplus \mathbb{T}^{\prime}_{k}$ \;
\textcolor{codegrey}{\tcc{tracklet flow}}
$\mathcal{F}_{k-1}$: \textcolor{codegreen}{$(\mathrm{B},\mathrm{d})$} $\leftarrow \mathbf{T}^{(L)}_{k-1}$ \;
\Return $\bar{\mathbb{T}}_{k}$: \textcolor{codegreen}{$(\mathrm{B}, \mathrm{10})$}, $\mathcal{F}_{k-1}$: \textcolor{codegreen}{$(\mathrm{B}, \mathrm{d})$}
\caption{Pseudo-code of VeloSSM-P}
\label{algo:velossm-p}
\end{algorithm}
%-------------------------------------------------------------------------
\begin{algorithm}[!t]
\SetAlgoLined
\DontPrintSemicolon
\SetNoFillComment
\footnotesize
\KwIn{tracklet velocity pred. \& obs. $\mathbb{\bar{T}}^{\prime}_{k},\mathbb{\tilde{D}}^{\prime}_{k}$: \textcolor{codegreen}{$(\mathrm{B}, \mathrm{1}, \mathrm{10})$};\\ \hspace{0.8cm}
tracklet previous state $\hat{\mathbb{T}}_{k-1}$: \textcolor{codegreen}{$(\mathrm{B}, \mathrm{1}, \mathrm{10})$};\\ \hspace{0.8cm}
tracklet obs. confidence $\mathbb{C}_{k}$: \textcolor{codegreen}{$(\mathrm{B}, \mathrm{1})$};\\ \hspace{0.8cm}
tracklet flow $\mathcal{F}_{k-1}$: \textcolor{codegreen}{$(\mathrm{B}, \mathrm{d})$};}
\KwOut{tracklet optimized state $\hat{\mathbb{T}}_{k}$: \textcolor{codegreen}{$(\mathrm{B}, \mathrm{10})$};}
\textcolor{codegrey}{\tcc{tracklet velocity gated embedding}}
$\mathbf{T}^{(0)}_{k}$: \textcolor{codegreen}{$(\mathrm{B}, \mathrm{1}, \mathrm{d})$} $\leftarrow \text{gMLP}_{\mathbb{C}_k}(\mathbb{\bar{T}}^{\prime}_{k},\mathbb{\tilde{D}}^{\prime}_{k})$ \;
\textcolor{codegrey}{\tcc{processing each layer of $L$ layers}}
\ForEach{$\ell \in \left[0, L \right)$}{
\textcolor{codegrey}{\tcc{normalize the input sequence}}
${\hat{\mathbf{T}}}^{(\ell)}_{k}$: \textcolor{codegreen}{$(\mathrm{B}, \mathrm{1}, \mathrm{d})$} $\leftarrow \text{RMSNorm}(\mathbf{T}^{(\ell)}_{k})$\;
\textcolor{codegrey}{\tcc{input projection}}
$\mathbf{x}$: \textcolor{codegreen}{$(\mathrm{B}, \mathrm{1}, \mathrm{D})$} $\leftarrow \text{SiLU}(\text{Conv1d}(\text{Linear}_{\mathbf{x}}({\hat{\mathbf{T}}}^{(\ell)}_{k})))$\;
$\mathbf{z}$: \textcolor{codegreen}{$(\mathrm{B}, \mathrm{1}, \mathrm{D})$} $\leftarrow \text{Linear}_{\mathbf{z}}({\hat{\mathbf{T}}}^{(\ell)}_{k})$\;
$\mathbf{A}$: \textcolor{codegreen}{$(\mathrm{D}, \mathrm{N})$} $\leftarrow  \text{Parameter}$ \;
\colorbox{Ocean}{\makebox[6.5cm][l]{\hspace*{-0.1cm}$\mathbf{B}$: \textcolor{codegreen}{$(\mathrm{B}, \mathrm{1}, \mathrm{N})$} $\leftarrow  \text{Linear}_{\mathbf{B}}(\mathcal{F}_{k-1})$}} \;\vspace{-2pt}
\colorbox{Ocean}{\makebox[6.5cm][l]{\hspace*{-0.1cm}$\mathbf{C}$: \textcolor{codegreen}{$(\mathrm{B}, \mathrm{1}, \mathrm{N})$} $\leftarrow  \text{Linear}_{\mathbf{C}}(\mathcal{F}_{k-1})$}} \;\vspace{-2pt}
\colorbox{Ocean}{\makebox[6.5cm][l]{\hspace*{-0.1cm}$\mathbf{\Delta}$: \textcolor{codegreen}{$(\mathrm{B}, \mathrm{1}, \mathrm{D})$} $\leftarrow \text{softplus}(\text{Broadcast}_{\mathrm{D}}(\text{Linear}_{\mathbf{\Delta}}(\mathcal{F}_{k-1})))$}} \; \vspace{-2pt}
\textcolor{codegrey}{\tcc{discretize}}
$\overline{\mathbf{A}}$: \textcolor{codegreen}{$(\mathrm{B}, \mathrm{1}, \mathrm{D}, \mathrm{N})$} $\leftarrow  \text{exp}(\mathbf{\Delta} \otimes  \mathbf{A})$ \;
$\overline{\mathbf{B}}$: \textcolor{codegreen}{$(\mathrm{B}, \mathrm{1}, \mathrm{D}, \mathrm{N})$} $\leftarrow \mathbf{\Delta} \otimes  \mathbf{B}$ \;
\textcolor{codegrey}{\tcc{Cross-SSM}}
$\mathbf{y}$: \textcolor{codegreen}{$(\mathrm{B}, \mathrm{1}, \mathrm{D})$} $\leftarrow \textbf{SSM}(\overline{\mathbf{A}}, \overline{\mathbf{B}}, \mathbf{C})(\mathbf{x})$\;
\textcolor{codegrey}{\tcc{gated output}}
$\mathbf{y}$: \textcolor{codegreen}{$(\mathrm{B}, \mathrm{1}, \mathrm{D})$} $\leftarrow \mathbf{y} \odot \text{SiLU}(\mathbf{z})$\;
\textcolor{codegrey}{\tcc{residual connection}}
$\mathbf{T}^{(\ell +1)}_{k}$: \textcolor{codegreen}{$(\mathrm{B}, \mathrm{1}, \mathrm{d})$} $\leftarrow \text{Linear}_{\mathbf{T}}(\mathbf{y})+\mathbf{T}^{(\ell)}_{k}$ \;
}
$\boldsymbol{w}_{k}$: \textcolor{codegreen}{$(\mathrm{B},\mathrm{10})$} $\leftarrow \text{sigmoid}(\text{Linear}_{\boldsymbol{w}}(\mathbf{T}^{(L)}_{k}))$ \;
$\mathbb{\hat{T}}^{\prime}_{k}$: \textcolor{codegreen}{$(\mathrm{B}, \mathrm{10})$} $\leftarrow \boldsymbol{w}_{k} \odot \mathbb{\tilde{D}}^{\prime}_{k} + (1-\boldsymbol{w}_{k}) \odot \mathbb{\bar{T}}^{\prime}_{k}$ \;
$\hat{\mathbb{T}}_{k}$: \textcolor{codegreen}{$(\mathrm{B}, \mathrm{10})$} $\leftarrow \hat{\mathbb{T}}_{k-1} \boxplus \mathbb{\hat{T}}^{\prime}_{k}$ \;
\Return $\hat{\mathbb{T}}_{k}$: \textcolor{codegreen}{$(\mathrm{B}, \mathrm{10})$}
\caption{Pseudo-code of VeloSSM-U}
\algorithmfootnote{The key steps of Cross-SSM are highlighted in blue.}
\label{algo:velossm-u}
\end{algorithm}
%-------------------------------------------------------------------------
\begin{algorithm}[!t]
\SetAlgoLined
\DontPrintSemicolon
\SetNoFillComment
\footnotesize
\KwIn{distance matrix $\mathbf{D}$: \textcolor{codegreen}{$(\mathrm{B},\mathrm{C},\mathrm{H},\mathrm{W})$};}
\KwOut{soft association matrix $\tilde{\mathbf{A}}$: \textcolor{codegreen}{$(\mathrm{B},\mathrm{H}, \mathrm{W})$};}
\textcolor{codegrey}{\tcc{patch embedding \& position encodings}}
$\mathbf{D}^{(0,0)}$: \textcolor{codegreen}{$(\mathrm{B}, \mathrm{d}, \mathrm{H}, \mathrm{W})$} $\leftarrow \text{LN}(\text{Conv2d}(\mathbf{D}))+ \mathbf{P}_{pos}$ \;
\textcolor{codegrey}{\tcc{processing each stage of $S$ stages}}
\ForEach{$s \in \left[0, S \right)$}{
\textcolor{codegrey}{\tcc{processing each level of stage $s$}}
\ForEach{$\ell \in \left[0, L_s \right)$}{
\textcolor{codegrey}{\tcc{input projection}}
$\hat{\mathbf{D}}^{(s,\ell)}$: \textcolor{codegreen}{$(\mathrm{B}, \mathrm{2^sD}, \mathrm{H}, \mathrm{W})$} $\leftarrow \text{Linear}(\text{LN}(\mathbf{D}^{(s,\ell)}))$\;
$\hat{\mathbf{D}}^{(s,\ell)}$: \textcolor{codegreen}{$(\mathrm{B}, \mathrm{2^sD}, \mathrm{H}, \mathrm{W})$} $\leftarrow \text{SiLU}(\text{DWConv}(\hat{\mathbf{D}}^{(s,\ell)}))$\;
\textcolor{codegrey}{\tcc{spatial-temporal bi-scan}}
$\mathbf{x}$: \textcolor{codegreen}{$(\mathrm{B},4,\mathrm{2^sD}, \mathrm{HW})$} $\leftarrow \textbf{Bi-Scan}(\mathbf{\hat{\mathbf{D}}^{(s,\ell)}})$\;
$\mathbf{A}$: \textcolor{codegreen}{$(\mathrm{4*2^sD}, \mathrm{N})$} $\leftarrow  \text{Parameter}$ \;
$\mathbf{B}$: \textcolor{codegreen}{$(\mathrm{B}, \mathrm{4}, \mathrm{N}, \mathrm{HW})$} $\leftarrow  \text{Linear}_{\mathbf{B}}(\mathbf{x})$ \;
$\mathbf{C}$: \textcolor{codegreen}{$(\mathrm{B}, \mathrm{4}, \mathrm{N}, \mathrm{HW})$} $\leftarrow  \text{Linear}_{\mathbf{C}}(\mathbf{x})$ \;
$\mathbf{\Delta}$: \textcolor{codegreen}{$(\mathrm{B}, \mathrm{4*2^sD}, \mathrm{HW})$} $\leftarrow \text{Linear}_{\mathbf{\Delta}}(\mathbf{\mathbf{x}})$ \;
\textcolor{codegrey}{\tcc{discretize}}
$\overline{\mathbf{A}}$: \textcolor{codegreen}{$(\mathrm{B}, \mathrm{4*2^sD}, \mathrm{HW}, \mathrm{N})$} $\leftarrow  \text{exp}(\mathbf{\Delta} \otimes  \mathbf{A})$ \;
$\overline{\mathbf{B}}$: \textcolor{codegreen}{$(\mathrm{B}, \mathrm{4*2^sD}, \mathrm{HW}, \mathrm{N})$} $\leftarrow \mathbf{\Delta} \otimes  \mathbf{B}$ \;
\textcolor{codegrey}{\tcc{SS2D}}
$\mathbf{y}$: \textcolor{codegreen}{$(\mathrm{B}, \mathrm{4*2^sD}, \mathrm{HW})$} $\leftarrow \textbf{SSM}(\overline{\mathbf{A}}, \overline{\mathbf{B}}, \mathbf{C})(\mathbf{x})$\;
\textcolor{codegrey}{\tcc{spatial-temporal bi-merge}}
$\mathbf{y}$:\scriptsize{\textcolor{codegreen}{$(\mathrm{B},\mathrm{2^sD},\mathrm{HW})$}} $\leftarrow \textbf{Bi-Merge}(\mathbf{y})$\;
\textcolor{codegrey}{\tcc{residual connection}}
$\mathbf{D}^{(s,\ell+1)}$: \textcolor{codegreen}{$(\mathrm{B}, \mathrm{2^sD}, \mathrm{H}, \mathrm{W})$} $\leftarrow \text{Linear}(\text{LN}(\mathbf{y})) + \mathbf{D}^{(s,\ell)}$\;
$\mathbf{D}^{(s,\ell+1)}$: \textcolor{codegreen}{$(\mathrm{B}, \mathrm{2^sD}, \mathrm{H}, \mathrm{W})$} $\leftarrow \text{LN}(\text{FFN}(\mathbf{D}^{(s,\ell+1)}))$\;
}
$\mathbf{D}^{(s+1,0)}$: \textcolor{codegreen}{$(\mathrm{B}, \mathrm{2^{s+1}D}, \mathrm{H}, \mathrm{W})$} $\leftarrow \text{LN}(\text{Conv2d}(\mathbf{D}^{(s,L_s)}))$ \;
}
$\tilde{\mathbf{A}}$: \textcolor{codegreen}{$(\mathrm{B},\mathrm{H}, \mathrm{W})$} $\leftarrow \text{sigmoid}(\text{Conv2d}(\text{LN}(\mathbf{D}^{(S,0)})))$ \;
\Return $\tilde{\mathbf{A}}$: \textcolor{codegreen}{$(\mathrm{B},\mathrm{H}, \mathrm{W})$}
\caption{Pseudo-code of HSSM}
\algorithmfootnote{$\mathrm{H}$ denotes the number of tracklets, $\mathrm{W}$ represents the number of detected objects, and LN stands for LayerNorm.}
\label{algo:hssm}
\end{algorithm}
%-------------------------------------------------------------------------
For implement, the appearance similarity $\mathbf{D}_{\text{App.}}$ is calculated using bi-directional softmax~\cite{pang2021quasi}, while the motion similarity is calculated using the motion-aware similarity~\cite{hu2022monocular},
\begin{align}
\mathbf{D}_{\text{App.}}(\mathbf{f}_{\mathbb{T}}, \mathbf{f}_{\mathbb{D}}) &= \frac{\sigma_x(\mathbf{f}_{\mathbb{T}} \mathbf{f}_{\mathbb{D}}^{\top}) + \sigma_y(\mathbf{f}_{\mathbb{T}} \mathbf{f}_{\mathbb{D}}^{\top})}{2}, \\
\mathbf{D}_{\text{mot.}}(\mathbf{v}_{\mathbb{T}}, \mathbf{v}_{\mathbb{D}}) &= \mathrm{w}_{\text{cos}}\mathbf{D}_{\text{cent.}} + (1-\mathrm{w}_{\text{cos}})\mathbf{D}_{pseu.}, \\
\mathbf{D}_{\text{cent.}}(\mathbf{B}_{\mathbb{T}}, \mathbf{B}_{\mathbb{D}}) &= \exp\left(-\frac{\|\mathbf{B}_{\mathbb{T}} - \mathbf{B}_{\mathbb{D}}\|_2}{10}\right), \\
\mathbf{D}_{\text{pseu.}}(\mathbf{v}_{\mathbb{T}}, \mathbf{v}_{\mathbb{D}}) &= \exp\left(-\frac{\|\mathbf{v}_{\mathbb{T}} - \mathbf{v}_{\mathbb{D}}\|_2}{5}\right)
\end{align}
where \(\sigma_x(\cdot)\) and \(\sigma_y(\cdot)\) denote sigmoid functions applied along the \(x\)- and \(y\)-axes, respectively. \(\mathrm{w}_{\text{cos}}\) is the normalized cosine similarity between the two motion vectors,
\begin{equation}
\mathrm{w}_{\text{cos}} = \frac{1}{2}(1 + \cos(\mathbf{v}_{\mathbb{T}}, \mathbf{v}_{\mathbb{D}}))
\end{equation}
%-------------------------------------------------------------------------
%\section{Additional Experiments}
%\label{sec:additional-HSSM-ablation}

%-------------------------------------------------------------------------
\section{More Visualization}
\label{sec:more-visualization}
\ifcompileimages
\begin{figure*}[ht]
\centering
\includegraphics[width=\textwidth]{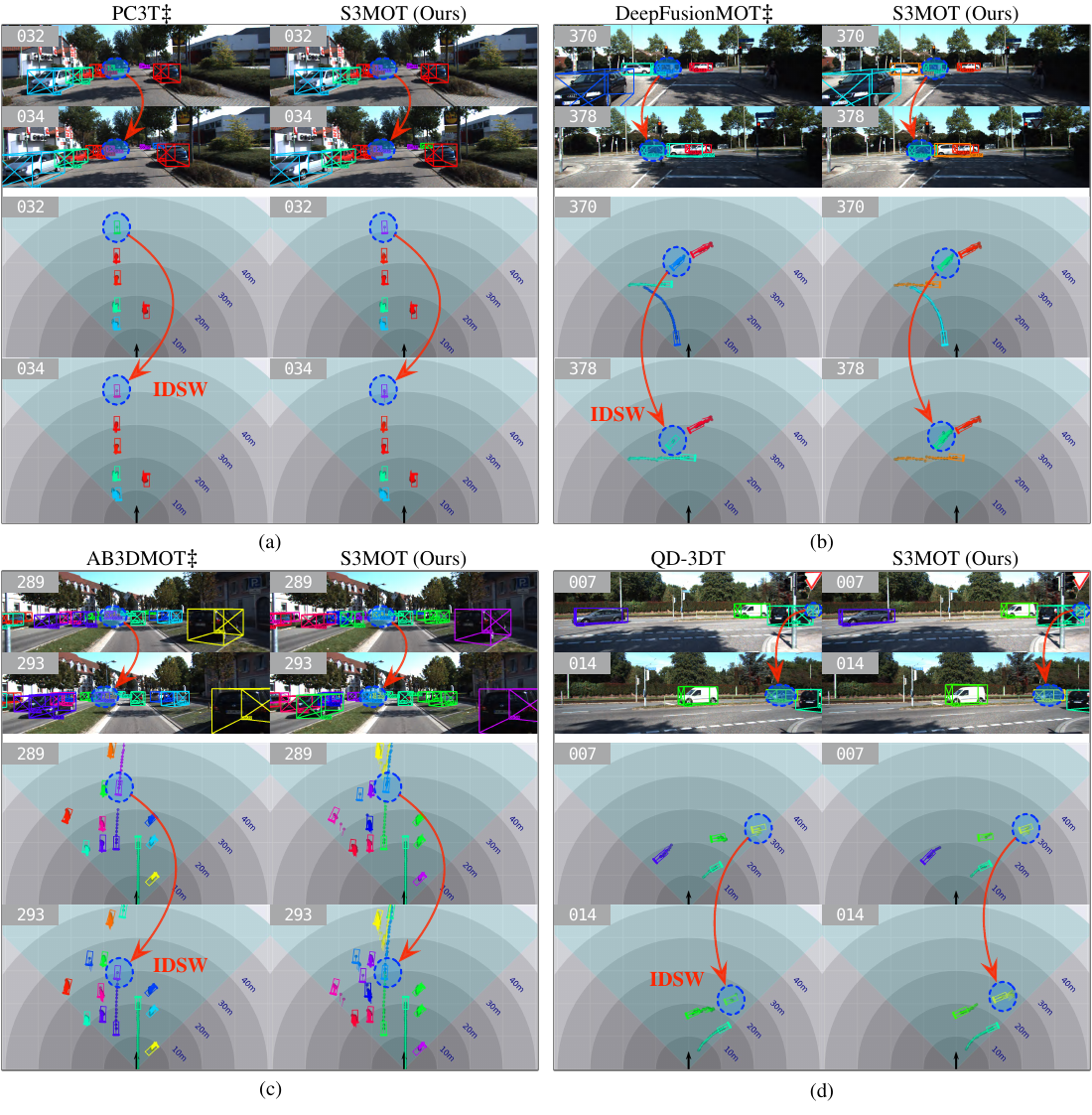}
\caption{Qualitative comparison between S3MOT and other state-of-the-art methods on the KITTI validation set. Methods marked with~$\ddag$ utilize the same DD3D detector as ours. Panels (a), (b), (c), and (d) highlight challenging scenarios where identity switching (IDSW) occurs when traditional linear assignment methods are used. S3MOT significantly mitigates IDSW through its data-driven data association strategy, achieving superior performance in maintaining consistent object identities across frames. Best viewed in color and zoom-in.}
\label{fg:s3mot_pc3t_ab3d}
\end{figure*}
\else
\textbf{Image compilation is disabled.}
\fi
\cref{fg:s3mot_pc3t_ab3d} demonstrates the effectiveness of S3MOT in addressing IDSW, a longstanding challenge in MOT. The figure showcases four challenging scenarios characterized by object occlusions, abrupt motion changes, and dense traffic. In comparison to existing methods such as PC3T~\cite{wu20213d}, DeepFusionMOT~\cite{wang2022deepfusionmot}, AB3DMOT~\cite{weng20203d}, and QD-3DT~\cite{hu2022monocular}, which often suffer from IDSW in these settings, S3MOT consistently delivers superior tracking performance. The frequent IDSW instances in the baseline methods highlight the inherent limitations of traditional linear assignment algorithms, such as the Hungarian Algorithm and Greedy Matching, particularly in scenarios involving overlapping objects or rapid trajectory shifts. By contrast, S3MOT utilizes a robust data-driven association framework that integrates spatial and temporal information, enabling it to maintain identity consistency across frames. This approach not only significantly reduces IDSW but also ensures reliable and accurate tracking in complex, real-world environments.

\end{document}